\documentclass[11pt]{article}

\usepackage[preprint]{acl}

\usepackage{times}
\usepackage{latexsym}

\usepackage[T1]{fontenc}

\usepackage[utf8]{inputenc}

\usepackage{microtype}

\usepackage{inconsolata}

\usepackage{graphicx}
\usepackage[utf8]{inputenc}
\usepackage[T1]{fontenc}
\usepackage{url}
\usepackage{booktabs}
\usepackage{amsfonts} 
\usepackage{amsmath}
\usepackage{amssymb}
\usepackage{nicefrac}
\usepackage{xcolor} 
\definecolor{Periwinkle}{RGB}{204,204,255}
\definecolor{mygrey}{gray}{0.6}
\usepackage{xspace}
\usepackage{pifont}
\usepackage{graphicx}
\usepackage{subcaption}
\usepackage{siunitx}
\usepackage{multirow}
\usepackage{enumitem}
\usepackage{makecell}
\usepackage{placeins}
\usepackage{colortbl}
\usepackage{algorithm}
\usepackage{algorithmicx}
\usepackage{algpseudocode}
\usepackage{caption}
\usepackage{tabularx}
\usepackage[table]{xcolor}
\usepackage{fvextra}

\usepackage{listings}
\usepackage{tcolorbox}
\tcbuselibrary{skins,breakable,listings}

\newcommand{\std}[1]{\scriptsize{$\pm$#1}}
\newcommand{\obsbox}[1]{%
    \begin{tcolorbox}[colframe=black!70, colback=lightgray!15, boxrule=1pt, arc=2mm]
        \small#1
    \end{tcolorbox}
}

\usepackage{hyperref}
\hypersetup{
    colorlinks=true,
    linkcolor=red,
    citecolor=cyan,
    filecolor=magenta,      
    urlcolor=magenta,
    }
\usepackage{soul}

\definecolor{White}{HTML}{ffffff}
\definecolor{DeepBlue}{HTML}{194a7a}
\definecolor{mygrey}{RGB}{200,200,200}
\definecolor{ErrorRed}{HTML}{8A3A3A}
\definecolor{WarningOrange}{HTML}{B07A3C}
\definecolor{NormalGreen}{HTML}{3A6F5D}
\definecolor{HighlightYellow}{HTML}{F2E6B8}
\definecolor{LightBlue}{HTML}{476f95}

\newcommand{\wrapcolorbox}[2]{%
  \sethlcolor{#1}\hl{#2}%
}
\title{Streaming Hallucination Detection in Long Chain-of-Thought Reasoning}

\author{%
\parbox{\linewidth}{\centering
{\bfseries
\makebox[\textwidth][c]{%
\begin{tabular}{@{}c@{\hspace{1.25em}}c@{\hspace{1.25em}}c@{\hspace{1.25em}}c@{\hspace{1.25em}}c@{}}
Haolang Lu\textsuperscript{1} &
Minghui Pan\textsuperscript{1} &
Ripeng Li\textsuperscript{1} &
Guoshun Nan\textsuperscript{1} &
Jialin Zhuang\textsuperscript{1}
\end{tabular}}\\
\makebox[\textwidth][c]{%
\begin{tabular}{@{}c@{\hspace{1.25em}}c@{\hspace{1.25em}}c@{\hspace{1.25em}}c@{\hspace{1.25em}}c@{}}
Zijie Zhao\textsuperscript{3} &
Zhongxiang Sun\textsuperscript{4} &
Kun Wang\textsuperscript{2} &
Yang Liu\textsuperscript{2} &
\end{tabular}}
}
\\[0.7ex]
{\normalfont
\textsuperscript{1}Beijing University of Posts and Telecommunications, China\\[0.8ex]
\textsuperscript{2}Nanyang Technological University, Singapore\\[0.8ex]
\textsuperscript{3}Southwest Jiaotong University, China\\[0.5em]
\textsuperscript{4}Renmin University of China, China\\[0.5em]
}
}
}

\begin{document}
\maketitle
\begin{abstract}
Long chain-of-thought (CoT) reasoning improves the performance of large language models, yet hallucinations in such settings often emerge subtly and propagate across reasoning steps.
We suggest that hallucination in long CoT reasoning is better understood as an evolving latent state rather than a one-off erroneous event.
Accordingly, we treat step-level hallucination judgments as local observations and introduce a cumulative prefix-level hallucination signal that tracks the global evolution of the reasoning state over the entire trajectory.
Overall, our approach enables streaming hallucination detection in long CoT reasoning, providing real-time, interpretable evidence.
\footnote{Code is available at \href{https://anonymous.4open.science/r/Streaming-Hallucination-Detection-D186/}{anonymous link}.}

\end{abstract}

\section{Introduction}

Reasoning large language model~\cite{dubey2024llama,wang2025large} have been widely adopted in complex tasks such as mathematical derivation~\cite{yu2025chain}, planning·~\cite{deng2025planu}, and multi-step question answering~\cite{sui2025fidelis}.
Compared to directly producing final answers, long chain-of-thought (CoT) reasoning~\cite{wei2022chain} improves performance by explicitly unfolding intermediate steps, making model outputs more interpretable and persuasive.
However, even within such fluent and seemingly well-justified reasoning processes, models may still arrive at incorrect conclusions, a failure mode commonly attributed to \textbf{HALLUCINATION}~\cite{zhang2025siren, huang2025survey}.

Hallucinations in long-CoT reasoning~\cite{cheng-etal-2025-chain} differ fundamentally from those in short reasoning or direct answer settings.
As reasoning unfolds over many steps, errors can emerge in increasingly subtle and diverse forms~\cite{lu2025auditing}.
In many cases, an incorrect final conclusion is accompanied by intermediate steps that are locally plausible and internally coherent, which mask the underlying error rather than correcting it and thus make hallucinations harder to identify.
To address this challenge, prior work has explored cross-model consistency checking~\cite{goel-etal-2025-zero,hou-etal-2025-probabilistic}, uncertainty-based confidence estimation~\cite{fadeeva-etal-2024-fact,NEURIPSQIU}, and analyses of internal model representations~\cite{su2024unsupervised,zhang2025prompt}.
These approaches have achieved notable progress in detecting hallucinations at the final-answer level or for isolated reasoning steps.

However, despite these advances, reliable hallucination detection in long CoT reasoning remains challenging.
This is largely because existing approaches lack an explicit characterization of how hallucinations \textbf{emerge} and \textbf{evolve}~\cite{YXLA2024-gsm2, sun2025detection} over extended reasoning trajectories~\cite{cheng-etal-2025-think,yao2025reasoning}.
Individual reasoning steps in long CoT are not independent decisions~\cite{YXLA2024-gsm1}, but instead jointly constitute a continuously evolving reasoning trajectory.
Consequently, hallucination judgments based on local evidence are often insufficient to determine whether the overall reasoning process has deviated from a factual foundation.
To better understand hallucination in long CoT reasoning and enable reliable streaming detection, we suggest that hallucination should be understood in a different way.

\vspace{-0.3em}
\begin{tcolorbox}[colframe=black!70, colback=lightgray!15, boxrule=1pt, arc=1mm]
        \includegraphics[width=0.4cm]{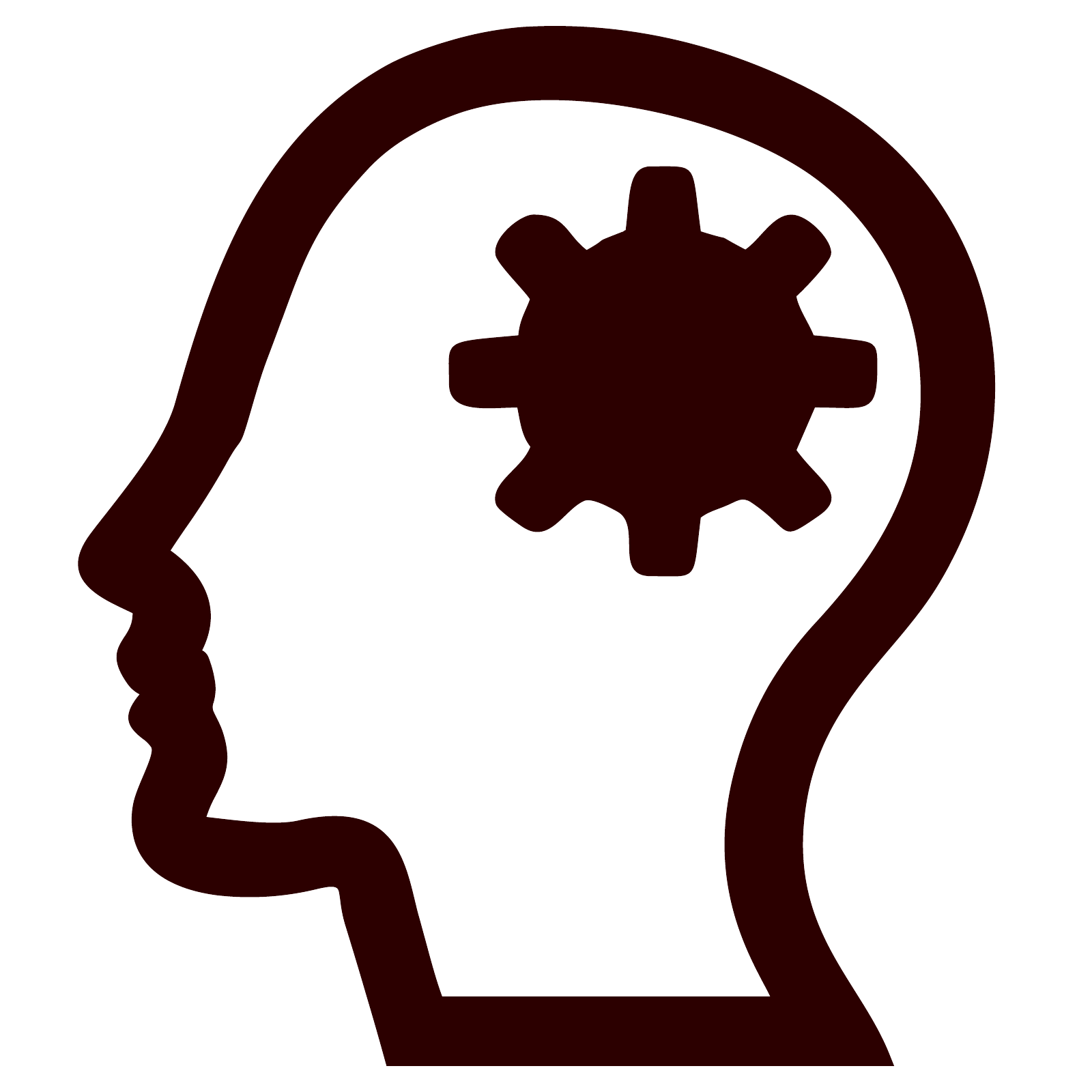}
        \textit{\small In long-CoT reasoning, hallucination is more appropriately understood not as a one-off erroneous event, but as a latent \textbf{state} that evolves over the course of the reasoning process.}
\end{tcolorbox}
\vspace{-0.3em}
This state reflects the extent to which a reasoning trajectory has come to be dominated by incorrect or insufficiently supported prefix~\cite{yu2025explainable}.
Rather than being fixed, this state is encoded in the model’s internal representations~\cite{ijcai2025p929,su2025activation} and is continuously updated as reasoning unfolds~\cite{marks2023geometry,zhang2025reasoning}, reflecting how the model revises its beliefs along the trajectory before settling on a final answer.
\textbf{This naturally yields a streaming formulation:} by continuously tracking the per-step hallucination state, we can thereby enable streaming modeling and detection of hallucination over long CoT trajectories.

\begin{figure}[t]
\centering
\includegraphics[width=0.45\textwidth]{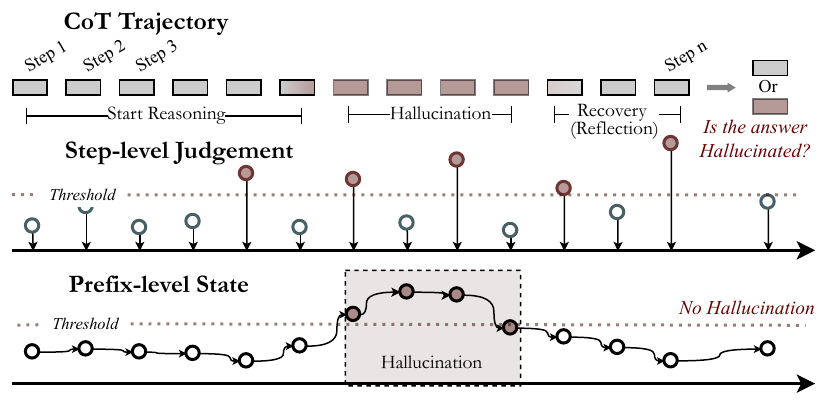}
\vspace{-2mm}
\caption{\footnotesize 
Hallucination as an evolving state in long CoT reasoning.
Step-level judgments capture local reasoning status at each reasoning step, while prefix-level hallucination represents the global state of the reasoning prefix.
}
\vspace{-7mm}
\label{fig:intro}
\end{figure}

As illustrated in Figure~\ref{fig:intro}, we characterize hallucination in long CoT reasoning through two signals along the reasoning trajectory.
At each reasoning step, we derive a \textbf{step-level hallucination} judgment as local evidence of whether the current step introduces unsupported content, which is obtained by probing the model's internal representations.
By nature, step-level hallucination can exhibit abrupt changes across adjacent steps along the trajectory, making step-level judgments alone insufficient to reliably determine whether the overall reasoning process has entered or remains in a hallucinated regime.
To further capture the global evolution of reasoning reliability, we further introduce a \textbf{prefix-level hallucination state} that integrates step-level evidence over time.
This prefix-level state represents whether the reasoning prefix has been persistently influenced by hallucinated premises rather than isolated local deviations.

In this work, we directly target the problem of \emph{streaming hallucination detection} and analyze \textbf{10{,}000+} long CoT trajectories with \textbf{200k+} reasoning steps, including \textbf{40k+} hallucinated steps.
\ding{68} We introduce a perspective that distinguishes \emph{step-level judgments} from \emph{prefix-level states}, under which hallucination signals become reliably decodable, achieving \textbf{over 87\% accuracy} at both levels.
\ding{68} Our approach enables \emph{online hallucination detection without additional inference cost}, correctly identifying \textbf{78\%} of CoT instances as reasoning unfolds; even when detection is imperfect, the resulting confidence trajectories provide \emph{stable and interpretable evidence} of the underlying reasoning state.
\ding{68} We further introduce \emph{eight logic-based dynamic metrics} and report \emph{seven empirical observations} that characterize how hallucination evidence propagates and recovers along the reasoning trajectory, offering actionable insights for future research on hallucination dynamics.




\section{Dataset \& Problem Definition}
\subsection{Hallucination as a Temporal Process} \label{subsec: problem define}
In this work, we model CoT reasoning as a prefix-conditioned temporal process.
Given an input $x$, a LLM
generates a sequence of reasoning steps 
{\abovedisplayskip=6pt
\belowdisplayskip=6pt
\begin{equation}
\mathcal{C}(x) = (s_1, s_2, \ldots, s_T), \qquad s_t \in \Sigma^{*},
\end{equation}}
where each $s_t$ denotes a contiguous segment of reasoning text.
Due to the autoregressive nature of LLMs, after generating the first $t$ steps, the model maintains an internal representation
{\abovedisplayskip=6pt
\belowdisplayskip=6pt
\begin{equation}
\mathbf{h}_t = \phi(x, s_{1:t}) \in \mathbb{R}^d,
\end{equation}}
which implicitly summarizes the entire reasoning prefix up to step $t$ and $d$ is the hidden dimension.

\paragraph{Step-level hallucination.}
We define \emph{step-level hallucination} as whether the current reasoning step introduces incorrect or unsupported information.
Let $z_t^{\mathrm{step}} \in \{0,1\}$ indicate the presence of hallucination evidence at step $t$.
The corresponding score is defined as
{\abovedisplayskip=6pt
\belowdisplayskip=6pt
\begin{equation}
c_t^{\mathrm{step}} \triangleq \mathbb{P}\!\left( z_t^{\mathrm{step}} = 1 \mid \mathbf{h}_t \right),
\end{equation}}
which serves as a local alarm signal.

\paragraph{Prefix-level hallucination state.}
To characterize the global reasoning state up to step $t$, we introduce a latent variable
$z_t^{\mathrm{prefix}} \in \{0,1\}$, indicating whether the reasoning prefix $s_{1:t}$ has entered a hallucinated state.
We define the \emph{prefix-level hallucination score} as
{\abovedisplayskip=6pt
\belowdisplayskip=6pt
{
\begin{equation}
c_t^{\mathrm{prefix}}
\triangleq 
\mathbb{P}\!\left(z_t^{\mathrm{prefix}}=1 \mid \mathbf{h}_t,\, c_t^{\mathrm{step}}\right)
\approx g_\theta\!\left(\mathbf{h}_t,\, c_t^{\mathrm{step}}\right),
\end{equation}}}
where $g_\theta(\cdot)$ denotes a lightweight predictor applied independently at each step.
Although inferred independently at each step, $c_t^{\mathrm{prefix}}$ provides a prefix-level assessment,
since $\mathbf{h}_t$ already encodes the entire reasoning trajectory.

At the final step $T$, $c_T^{\mathrm{prefix}}$ yields an overall judgment of whether the full CoT reasoning, and consequently its final answer, are hallucinated.

\subsection{Dataset Construction}
The dataset is annotated following the prefix-level hallucination introduced in Section~\ref{subsec: problem define}.The queries in the dataset are derived from existing work~\cite{trivedi-etal-2022-musique,suzgun2022challenging}.

\paragraph{Annotation targets.}
Two types of labels are assigned by \texttt{Claude~4.5} along each reasoning trajectory.
For each reasoning step $s_t$, we annotate a binary \emph{step-level hallucination label} $A_t^{\mathrm{step}}$.
and a \emph{prefix-level hallucination label} $A_t^{\mathrm{prefix}}$.
These labels correspond to the step-level judgment $c_t^{\mathrm{step}}$ and the prefix-level state $c_t^{\mathrm{prefix}}$, respectively.

\paragraph{Annotation validation.}
In addition to automatic labeling, we apply the following validation procedures:
\ding{202} \emph{Answer-aware semantic check}: global answer correctness is determined by a semantic judge to guide the annotation, ensuring alignment between reasoning chains and final outcomes;
\ding{203} \emph{Logical consistency check}: samples are filtered based on terminal constraints and state transition rules to eliminate logical paradoxes (e.g., severe logical epiphany);
\ding{204} \emph{Manual verification}: human experts review a stratified sample of the cleaned data to validate fine-grained step labels and correction logic.
Detailed annotation and validation procedures are provided in the Appendix~\ref{appx:data_validation}.

\begin{table}[t]
\centering
\resizebox{\columnwidth}{!}{%
\begin{tabular}{l|ccc}
\toprule
\textbf{Statistic} 
& \textbf{LLaMA} 
& \textbf{Qwen} 
& \textbf{DeepSeek} \\
\midrule
Total samples (questions) 
& 3{,}400 
& 3{,}000 
& 3{,}500 \\
Filtered usable samples 
& $\sim$2{,}500 
& $\sim$2{,}900 
& $\sim$2{,}800 \\
Final-answer hallucination rate 
& 73.47\% 
& 66.63\% 
& 48.33\% \\
\midrule
Total reasoning steps 
& 58{,}619 
& 53{,}728  
& 89{,}918 \\
Step-level hallucination rate 
& 38.80\% 
& 36.04\% 
& 13.74\% \\
Prefix-level hallucination rate 
& 58.96\% 
& 50.73\% 
& 27.64\% \\
\midrule
Average steps per CoT 
& 23.32 
& 18.20 
& 32.02 \\
\bottomrule
\end{tabular}
}
\label{tab:dataset_statistics}
\caption{\footnotesize Statistics of long-CoT hallucination dataset.
Annotations and validation are performed using multiple base models, including \texttt{Qwen}, \texttt{LLaMA}, and \texttt{DeepSeek}.
Due to model-dependent filtering and validation outcomes, the exact number of usable samples differs across models.}
\vspace{-1.2em} %
\end{table}

\begin{figure*}[t]
    \centering
    \includegraphics[width=1.0\textwidth]{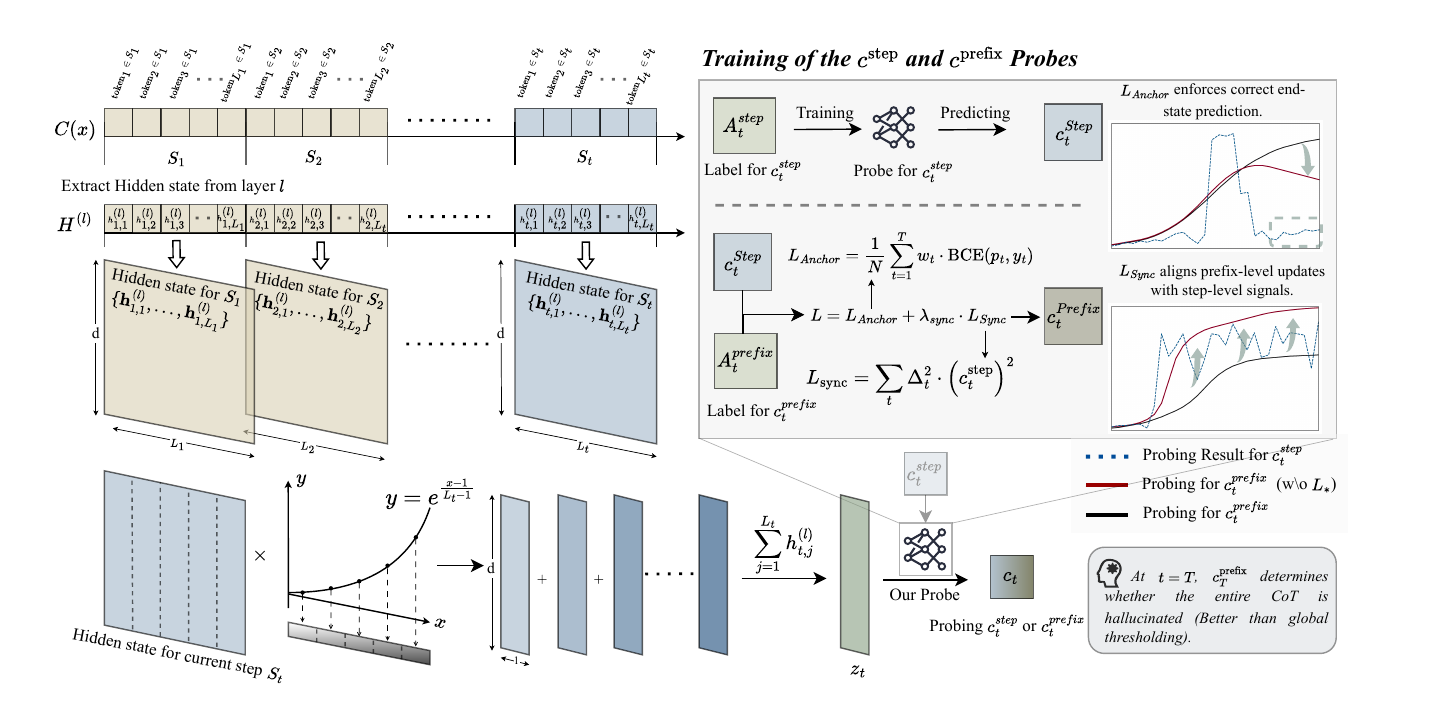}
    \vspace{-2.5em}
    \caption{\footnotesize Overview of the probing framework.
Both the step-level and prefix-level probes take the same step representation $\mathbf{z}_t$ as input.
While the step-level probe is trained to predict the label $A_t^{\mathrm{step}}$, the prefix-level probe targets global reasoning state $c_t^{\mathrm{prefix}}$.
To capture accumulated hallucination effects along the trajectory, the prefix-level probe is trained with prefix-level supervision $A_t^{\mathrm{prefix}}$, while using the step-level signal $c_t^{\mathrm{step}}$ as a guiding signal to facilitate accurate estimation of $c_t^{\mathrm{prefix}}$.
}
    \label{fig: overview}
    \vspace{-1.2em} %
\end{figure*}

\section{Step-level Hallucination Confidence}

In this section, we investigate step-level hallucination estimation for $A_t^{\mathrm{step}}$, propose a probe-based method, and validate it through empirical analysis.

\subsection{Probing Step-level Hallucination Signals} \label{subsec: step motivation}

Step-level hallucination reflects whether incorrect or unsupported information is introduced at a specific reasoning step.
To estimate such signals from LLMs, a natural approach is to apply a lightweight probe~\cite{tenneyyou, AL2023knowledge1} to the model’s hidden representations.
However, in autoregressive language models, hidden representations are strongly conditioned on long prefixes.
As a result, the effectiveness of probing critically depends on how step-level representations are constructed from token-level states.

Specifically, given a reasoning trajectory $C(x)$, each step $s_t$ is composed of $L_t$ tokens, where $L_t$ may vary across steps.
The hidden state of the $j$-th token in step $s_t$ at layer $l$ is denoted by $\mathbf{h}^{(l)}_{t,j} \in \mathbb{R}^d$, where $t$ indexes the step
(In this work, a sentence is treated as one reasoning step.).
Collectively, the token-level hidden states at layer $l$ for the reasoning prefix up to step $t$ form
$\mathbf{H}^{(l)} \in \mathbb{R}^{(\sum_{t} L_t)\times d}.$

To obtain a step-level representation at reasoning step $t$, a natural approach~\cite{liu2019linguistic, belinkov2022probing, zhang2025reasoning} is to construct a vector
$\mathbf{z}_t \in \mathbb{R}^d$
by aggregating token-level hidden states.
In practice, this aggregation is almost exclusively additive, taking the form
{\abovedisplayskip=6pt
\belowdisplayskip=6pt
\begin{equation}
\mathbf{z}_t
=
\sum_{i=1}^{t}
\sum_{j=1}^{L_i}
\mathbf{h}^{(l)}_{i,j} \in \mathbb{R}^d,
\end{equation}}
or its normalized variant.
Such constructions are intuitive, as they aim to summarize the model’s internal state up to the current reasoning position.
However, we observe two systematic issues with additive aggregation schemes that directly affect step-level hallucination estimation. 

\ding{68} \textbf{Property I (Cross-step saturation).}
When $\mathbf{z}_t$ is formed by additively aggregating all token hidden states up to step $s_t$, the incremental contribution of each new step diminishes as the prefix grows.
As a result, later-step representations saturate and become hard to separate for step-level classification.

\ding{68} \textbf{Property II (Within-step imbalance).}
Within a reasoning step, later tokens are conditioned on earlier ones and tend to capture a more complete summary of the step.
Uniformly aggregating token representations treats all positions equally, assigning smaller effective weights to later tokens and underemphasizing their contributions~\cite{zhou2025length, dong2021attention}.

(The detailed discussion is in Appendix~\ref{appx:theoretical_analysis}.)
\vspace{-0.6em} %
\subsection{Step-level Confidence Estimation}
\label{sec: step-level}

To satisfy Property~I and II, we construct step-level representations by aggregating token hidden states \emph{only within the current step} using a fixed, time-aware exponential weighting.
Specifically, for a reasoning step $s_t$ consisting of $L_t$ tokens with hidden states $\{\mathbf{h}^{(l)}_{t,1}, \ldots, \mathbf{h}^{(l)}_{t,L_t}\}$, we compute an unnormalized step representation as
{\abovedisplayskip=6pt
\belowdisplayskip=6pt
\begin{equation}
\tilde{\mathbf{z}}_t
=
\sum_{j=1}^{L_t}
\frac{\exp(w_j)}{\sum_{k=1}^{L_t} \exp(w_k)}
\,\mathbf{h}^{(l)}_{t,j},
\qquad
w_n = \frac{n-1}{L_t-1}.
\end{equation}
}
The final step-level representation is obtained by applying $\ell_2$ normalization,
{\abovedisplayskip=4pt
\belowdisplayskip=4pt
\begin{equation}
\mathbf{z}_t
=
\frac{\tilde{\mathbf{z}}_t}{\lVert \tilde{\mathbf{z}}_t \rVert_2}
\in \mathbb{R}^d.
\end{equation}
}

As illustrated in Figure~\ref{fig: overview}, the resulting vector $\mathbf{z}_t$ serves as the input to a step-level probe.
The probe is trained using the step-level hallucination label $A_t^{\mathrm{step}}$ and is applied to hidden states from a specific layer $l$, enabling targeted probing of step-level hallucination signals.
This construction assigns exponentially larger weights to later tokens within a step, emphasizing information introduced toward the end of the step while avoiding aggregation over long prefixes.
(Other aggregation schemes, including uniform averaging and linear weighting, are treated as baselines in our experiments.)

\vspace{-0.5em}
\subsection{Key Findings}
\vspace{-0.6em}
In this subsection, we present empirical results and key observations from our investigation of step-level hallucination in long-CoT reasoning.
We evaluate step-level probes across different base models, analyze their sensitivity along the reasoning trajectory, and examine how hallucination signals are distributed across layers.
\vspace*{-1em}
\paragraph{Cross-Model Comparison of Step-level Probes.}
Table~\ref{tab: step} reports step-level probing performance.
Under identical step-level supervision, our method consistently outperforms all baselines across models and metrics.
Compared with the global-mean aggregation baseline, our representation improves AUC by approximately $+4.6\%$, $+5.8\%$, and $+2.9\%$ on three base models with corresponding gains in ACC and F1.

\vspace{-0.3em}
\obsbox
{\textbf{\textit{Obs I.}} Under identical supervision, step-level representation construction alone leads to consistent and measurable performance gains.}
\vspace{-0.3em}

Since all methods are trained with the same step-level labels $A_t^{\mathrm{step}}$, these improvements cannot be attributed to supervision strength or probe capacity.
Instead, the results indicate that modifying the probe input representation mitigates cross-step saturation and within-step imbalance, providing support for \textbf{Property~I} and \textbf{Property~II}.

\begin{figure}[h]
\centering
\includegraphics[width=0.45\textwidth]{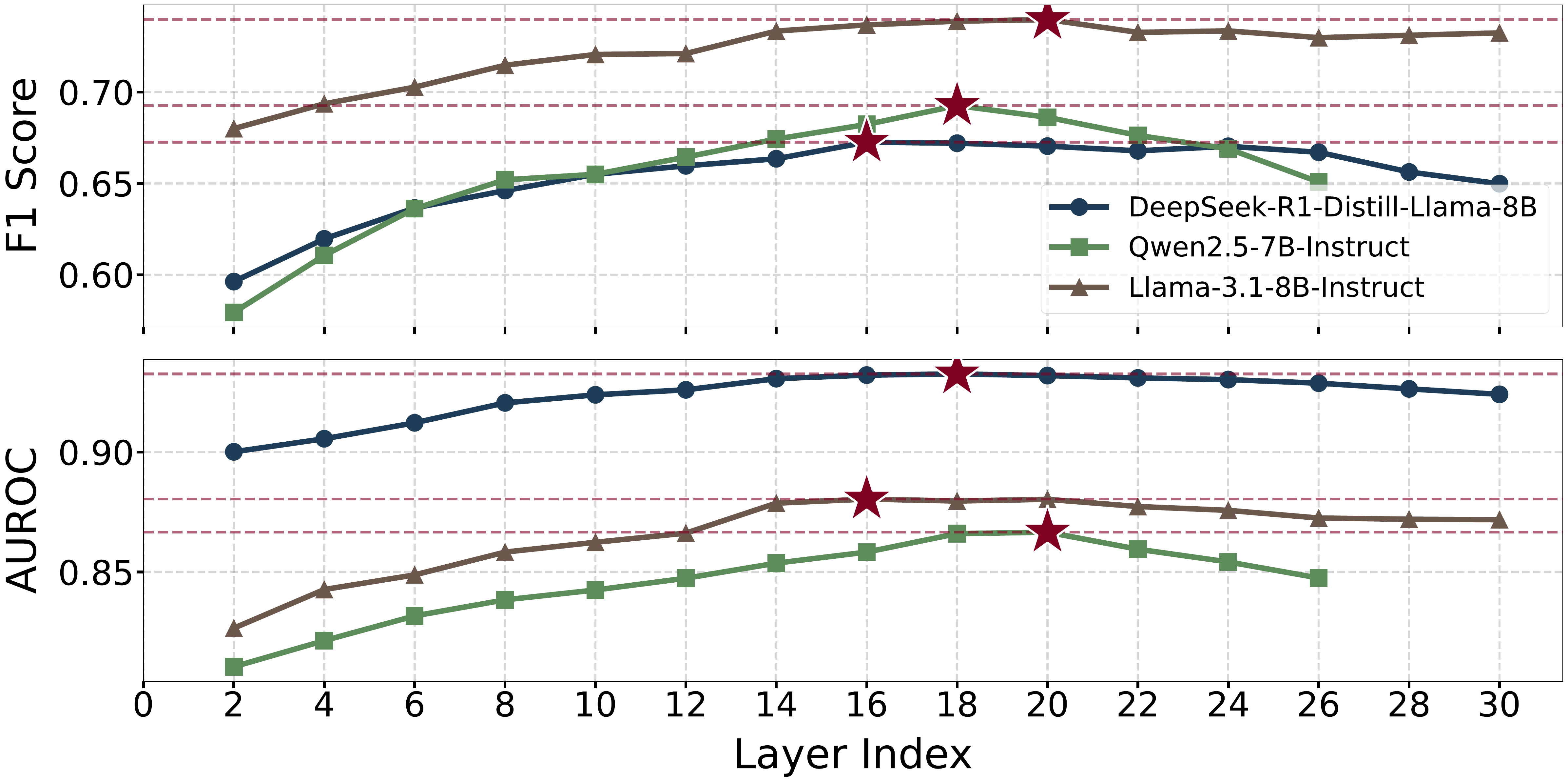}
\vspace{-1em}
\caption{\footnotesize 
Layer-wise analysis of step-level hallucination probing.
F1 score (top) and AUROC (bottom) are reported across transformer layers for three base models.
}
\label{fig:layers}
\vspace{-1.5em} %
\end{figure}

\begin{table*}[t]
\centering

\large
\renewcommand{\arraystretch}{1.15}
\resizebox{\textwidth}{!}{
\begin{tabular}{c|ccc|ccc|ccc}
\toprule[1.5pt]
\raisebox{-1.2ex}{\textbf{Method}} &
\multicolumn{3}{c|}{\textbf{Llama-3.1-8B-Instruct}} &
\multicolumn{3}{c|}{\textbf{Qwen2.5-7B-Instruct}} &
\multicolumn{3}{c}{\textbf{DeepSeek-R1-Distill-8B}} \\
\cmidrule(lr){2-4} \cmidrule(lr){5-7} \cmidrule(lr){8-10}
& \textbf{AUC} & \textbf{ACC} & \textbf{F1}
& \textbf{AUC} & \textbf{ACC} & \textbf{F1}
& \textbf{AUC} & \textbf{ACC} & \textbf{F1} \\
\midrule

TTPD & {60.44\%\std{1.07\%}} & {59.81\%\std{1.49\%}} & {68.51\%\std{1.07\%}} & {57.51\%\std{1.06\%}} & {75.06\%\std{2.17\%}} & {62.24\%\std{2.07\%}} & {68.32\%\std{2.17\%}} & {48.91\%\std{3.52\%}} & {55.22\%\std{3.61\%}} \\
SAPLMA & {82.83\%\std{1.21\%}} & {76.01\%\std{1.05\%}} & {65.64\%\std{1.24\%}} & {82.04\%\std{2.03\%}} & {74.24\%\std{1.80\%}} & {62.73\%\std{1.38\%}} & {90.93\%\std{1.67\%}} & {91.64\%\std{1.48\%}} & {63.93\%\std{2.04\%}} \\
Global Mean & {82.27\%\std{1.43\%}} & {75.20\%\std{1.17\%}} & {64.26\%\std{1.83\%}} & {80.95\%\std{1.25\%}} & {73.50\%\std{1.76\%}} & {61.38\%\std{2.12\%}} & {90.38\%\std{1.25\%}} & {91.19\%\std{1.62\%}} & {62.24\%\std{1.69\%}} \\
\textbf{Ours} &
\textbf{87.83\%\std{1.60\%}} & \textbf{80.37\%\std{1.97\%}} & \textbf{72.42\%\std{1.03\%}} &
\textbf{86.70\%\std{1.92\%}} & \textbf{78.65\%\std{1.04\%}} & \textbf{69.65\%\std{1.37\%}} &
\textbf{93.27\%\std{2.19\%}} & \textbf{92.74\%\std{1.82\%}} & \textbf{67.27\%\std{2.13\%}} \\

\bottomrule[1.5pt]
\end{tabular}
}
\caption{\footnotesize \textbf{Empirical comparison of step-level hallucination probing methods.} We compare two representative baselines~\cite{burger2024truth, azaria2023internal} that perform error detection using internal model representations.
We additionally include a global-mean aggregation baseline, corresponding to the unmodified probe input construction in Section~\ref{subsec: step motivation}.
To ensure experimental fairness, all methods are trained under the same step-level supervision using labels $A_t^{\mathrm{step}}$.
}
\label{tab: step}
\vspace{-1.2em} %
\end{table*}

\vspace*{-0.1em}
\paragraph{Layer-wise Analysis of Hallucination Probing}
From Figure~\ref{fig:layers}, we observe that step-level hallucination probing achieves the strongest performance at intermediate layers across models.
Interestingly, this observation differs from prior work that probes individual token representations, where shallow-layer features typically perform poorly.
In contrast, we find that the performance gap across layers—including very shallow ones—is relatively small, which we attribute to aggregating multiple token representations within each step.
This suggests that later tokens inherit reasoning-relevant information formed by earlier tokens at deeper layers, providing indirect support for \textbf{Property~II}.

\vspace{-0.2em}
\paragraph{Temporal Sensitivity along the CoT.}

\begin{figure}[h]
\centering
\includegraphics[width=0.48\textwidth]{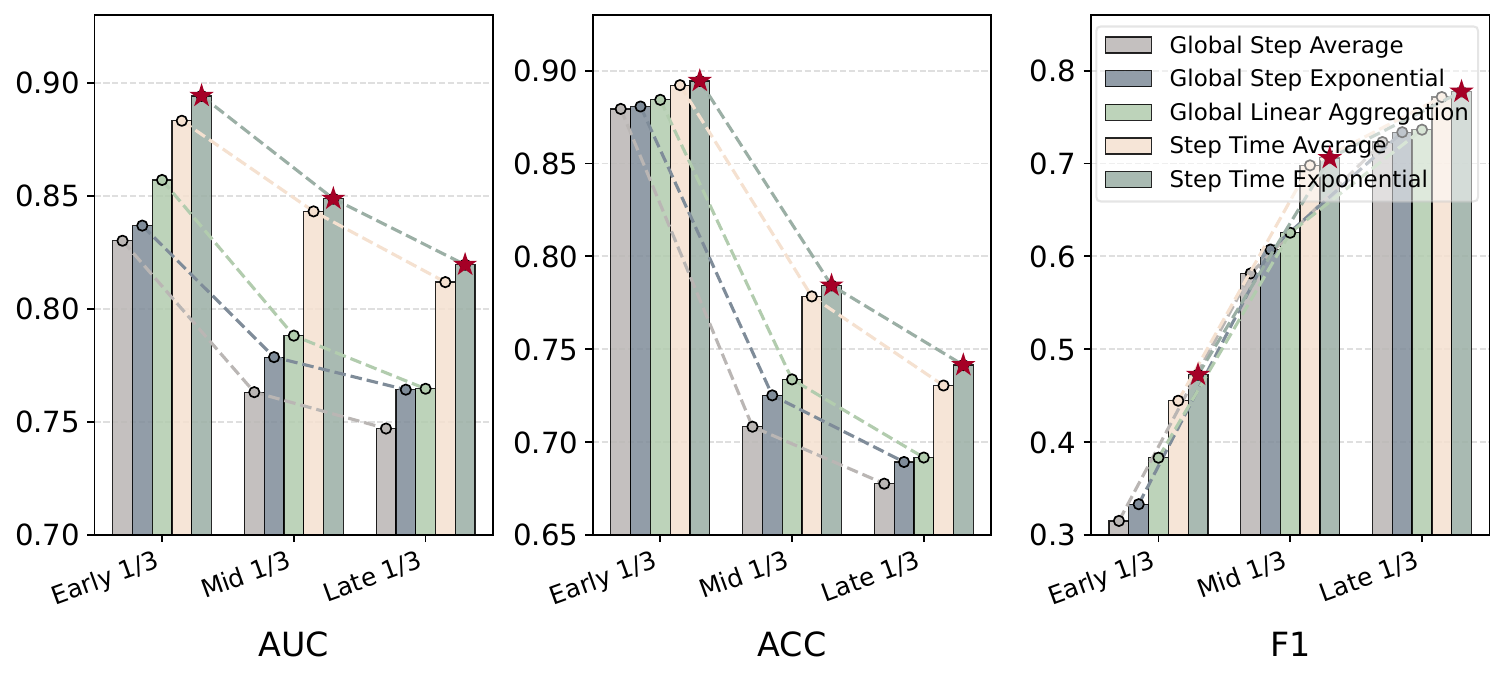}
\vspace{-2em}
\caption{\footnotesize 
Step-level probing performance across different CoT positions on \texttt{LLaMA-3.1-8B}.
We compare five probe variants with different representation aggregation strategies, where \emph{Step Time Exponential} is our final choice.
AUC is threshold-free, while ACC and F1 use a fixed threshold of $0.5$.
}
\label{fig:3stage}
\vspace{-0.8em} %
\end{figure}

To examine how step-level probing performance varies along the reasoning trajectory, we evaluate probes at different CoT positions by grouping steps into early, middle, and late thirds in Figure~\ref{fig:3stage}.
Across all probe variants, AUC decreases from approximately $0.89$ in early steps to about $0.82$ in late steps, and accuracy drops from around $0.89$ to $0.74$.
In contrast, F1 increases from below $0.35$ in early steps to nearly $0.78$ in later stages.

A direct interpretation of this trend is that the declining AUC and accuracy align with our earlier hypothesis that step-level representations become less discriminative in later stages of long-CoT reasoning.
As the chain grows, newly introduced errors and normal steps are increasingly embedded in accumulated context, making them harder to distinguish at the representation level.

\obsbox{
\textbf{Obs 2.} As reasoning progresses, step-level representations become less separable for newly introduced errors versus normal steps.

}

To further understand the contrasting behavior of F1, we analyze its precision and recall.
While precision remains stable across the CoT (approximately $0.74$--$0.75$), recall increases markedly, from about $0.31$--$0.34$ in early steps to over $0.80$ in late steps, indicating that the rise in F1 is primarily recall-driven.
We attribute this pattern to the combined effect of label density shift (with fewer errors early on) and evidence strength along the reasoning trajectory.
However, the fact that our step-level representation yields measurable gains at early stages indicates that hallucination evidence is not inherently uncapturable early on, but is simply more difficult to detect than at later stages.

\obsbox{\textbf{Obs 3.} Early hallucinations are detectable at the step level but exhibit sparse and weak evidence, whereas later steps accumulate stronger signals, leading to persistent errors even under step-level probing.}


\section{Prefix-level Hallucination State} \label{sec: prefix}
In this section, we model prefix-level hallucination to characterize whether the reasoning process up to a given step has entered a hallucinated state, and introduce a step-guided estimation method with a comprehensive evaluation.

\subsection{Step-wise Evidence to Prefix-level State}

Following the definition of prefix-level hallucination in Eq.~(4), the prefix-level score is not only conditioned on the hidden representation $\mathbf{h}_t$, but also explicitly incorporates the step-level signal $c_t^{\mathrm{step}}$, which captures newly introduced local evidence.
This design reflects the fact that, although $\mathbf{h}_t$ encodes the entire reasoning prefix, hallucination evidence is typically introduced incrementally at each reasoning step.

Formally, the transition of hidden representations between consecutive steps can be expressed as
$\mathbf{h}_{t+1} = \mathbf{h}_t + \mathbf{u}_{t+1},$
where $\mathbf{u}_{t+1}$ denotes the state change induced by the current step.
Therefore, the step-level signal $c_{t+1}^{\mathrm{step}}$ can be viewed as an estimator of the newly introduced state increment $\mathbf{u}_{t+1}$, rather than of the global reasoning state itself.
Consequently, incorporating $c_t^{\mathrm{step}}$ into the computation of $c_t^{\mathrm{prefix}}$ is essential for aligning prefix-level estimation with step-wise state changes, and for recovering a hallucination signal that meaningfully reflects the underlying reasoning dynamics.

However, since $c_t^{\mathrm{prefix}}$ is intended to represent a latent reasoning state rather than an instantaneous alarm, its evolution should satisfy additional structural constraints.
In particular, we identify two desiderata that a well-formed prefix-level hallucination score should meet.


\ding{68} \textbf{Property~III (Temporal coherence).}
The prefix-level hallucination score represents a latent reasoning state
and should evolve in a temporally coherent manner along the reasoning trajectory.
Specifically, the score should vary smoothly in general,
while allowing sharp transitions when sufficient step-level evidence is introduced.

\ding{68} \textbf{Property~IV (Directional consistency).}
The prefix-level hallucination score should remain responsive to step-level evidence and be allowed to both increase and decrease over time, enabling transitions toward either a hallucinated or a correct final reasoning state.

Taken together, these properties emphasize that prefix-level hallucination modeling is not a simple accumulation of step-level alarms.
Instead, it requires integrating local, potentially noisy estimates of state increments into a temporally coherent trajectory that ultimately aligns with the correctness of the final reasoning outcome.

\vspace{-0.3em} %
\subsection{Step-guided Prefix-level Estimation}

Following \textbf{Properties~III and~IV}, our objective is to learn a prefix-level hallucination predictor that fits the supervision signal $A_t^{\mathrm{prefix}}$ while maintaining temporal continuity and directional consistency.
Specifically, the prefix-level score should evolve smoothly over time and remain responsive to step-level hallucination evidence, without enforcing irreversible accumulation.

To achieve this, we explicitly incorporate the step-level hallucination score $c_t^{\mathrm{step}}$ during both training and inference as a source of local evidence, rather than as an additional label.
We adopt a logic-enhanced training strategy with two complementary objectives: a final anchor that enforces correctness at the end of the reasoning trajectory, and a step-guided synchronization term that reduces detection lag while preserving the model’s ability to recover through later reasoning.

We impose a strong supervision signal at the final reasoning step using a weighted binary cross-entropy loss,
{\abovedisplayskip=6pt
\belowdisplayskip=6pt
\begin{equation}
\begin{aligned}
L_{\mathrm{anchor}}
&= \frac{1}{T}
\sum_{t=1}^{T}
w_t \cdot \mathrm{BCE}(c_t^{\mathrm{prefix}}, A_t^{\mathrm{prefix}}), \\
w_t
&=
\begin{cases}
\lambda_{\mathrm{final}}, & t = T, \\
1, & \text{otherwise}.
\end{cases}
\end{aligned}
\end{equation}}

This anchor loss counteracts long-horizon inertia and forces the prefix-level prediction to converge to the correct final state.

To incorporate step-level evidence without constraining recovery, we introduce a quadratic alarm synchronization loss.
Let $c_t^{\mathrm{prefix}}$ and $c_t^{\mathrm{step}}$ denote the prefix-level and step-level hallucination scores, respectively.
We define a one-way discrepancy
{\abovedisplayskip=6pt
\belowdisplayskip=6pt
\begin{equation}
\Delta_t = \max\!\left(0,\; c_t^{\mathrm{step}} - c_t^{\mathrm{prefix}} \right),
\end{equation}}
preix and the synchronization loss as
{\abovedisplayskip=6pt
\belowdisplayskip=6pt
\begin{equation}
L_{\mathrm{sync}} =
\sum_t
\Delta_t^2 \cdot \left(c_t^{\mathrm{step}}\right)^2 .
\end{equation}}

This loss penalizes missed alarms while suppressing low-confidence noise, enforcing directional consistency without inducing monotonic accumulation.
The overall training objective is
{\abovedisplayskip=6pt
\belowdisplayskip=6pt
\begin{equation}
L = L_{\mathrm{anchor}} + \lambda_{\mathrm{sync}} \, L_{\mathrm{sync}},
\end{equation}}
which implements the desired prefix-level behavior by jointly enforcing end-state correctness and locally guided, temporally coherent updates.

\begin{table*}[t]
\centering

\vspace{-0.8em}
\large
\renewcommand{\arraystretch}{1.15}
\resizebox{\textwidth}{!}{
\begin{tabular}{cc|ccc|ccc|ccc}
\toprule[1.5pt]
\raisebox{-1.2ex}{} &
\raisebox{-1.2ex}{\textbf{Method}} &
\multicolumn{3}{c|}{\textbf{Llama-3.1-8B-Instruct}} &
\multicolumn{3}{c|}{\textbf{Qwen2.5-7B-Instruct}} &
\multicolumn{3}{c}{\textbf{DeepSeek-R1-Llama-8B}} \\
\cmidrule(lr){3-5} \cmidrule(lr){6-8} \cmidrule(lr){9-11}
& &
\textbf{AUC} & \textbf{ACC} & \textbf{F1} &
\textbf{AUC} & \textbf{ACC} & \textbf{F1} &
\textbf{AUC} & \textbf{ACC} & \textbf{F1} \\
\midrule

\multirow{4}{*}{\rotatebox{90}{\textbf{Local}}} &
ICR & 81.06\%\std{1.34\%} & 73.45\%\std{2.17\%} & 78.09\%\std{1.91\%} & 75.07\%\std{2.48\%} & 68.68\%\std{1.05\%} & 72.31\%\std{1.72\%} &  83.82\%\std{1.34\%} & 85.15\%\std{2.06\%} & 62.91\%\std{2.13\%} \\
& LLM-Check & 80.16\%\std{1.83\%} & 74.08\%\std{2.12\%} & 76.50\%\std{1.14\%} & 80.07\%\std{2.36\%} & 73.57\%\std{1.59\%} & 75.45\%\std{0.94\%} & 70.89\%\std{1.12\%} & 63.62\%\std{2.09\%} & 47.21\%\std{2.14\%} \\
& Global Mean & 86.71\%\std{2.19\%} & 79.40\%\std{1.47\%} & 82.66\%\std{1.73\%} & 86.72\%\std{0.97\%} & 77.64\%\std{2.23\%} & 79.52\%\std{1.88\%}& 87.28\%\std{1.92\%} & 87.37\%\std{2.17\%} & 70.83\%\std{1.54\%} \\
& \textbf{Ours} &
\textbf{87.30\%\std{1.62\%}} & \textbf{79.38\%\std{2.31\%}} & \textbf{82.45\%\std{1.07\%}} &
\textbf{88.02\%\std{1.45\%}} & \textbf{78.90\%\std{0.92\%}} & \textbf{80.48\%\std{2.28\%}} &
\textbf{87.98\%\std{1.41\%}} & \textbf{87.14\%\std{1.75\%}} & \textbf{69.43\%\std{2.16\%}} \\

\midrule[1pt]

\multirow{4}{*}{\rotatebox{90}{\textbf{Final}}} &
ICR & 57.40\%\std{1.86\%} & 74.05\%\std{1.33\%} & 84.52\%\std{2.21\%} & 55.89\%\std{2.44\%} & 65.35\%\std{0.91\%} & 78.53\%\std{1.73\%} & 85.58\%\std{2.79\%} & 77.40\%\std{1.76\%} & 71.78\%\std{1.65\%}  \\
& LLM-Check & 56.85\%\std{2.15\%} & 77.27\%\std{1.09\%} & 87.18\%\std{1.57\%} & 65.10\%\std{2.38\%} & 69.30\%\std{1.68\%} & 81.79\%\std{0.93\%} & 66.10\%\std{1.12\%} & 56.47\%\std{2.43\%} & 60.50\%\std{1.56\%} \\
& Global Mean & 61.14\%\std{1.02\%} & 77.65\%\std{2.16\%} & 87.09\%\std{1.77\%} & 78.70\%\std{1.21\%} & 71.27\%\std{2.05\%} & 82.65\%\std{1.46\%}& 91.34\%\std{0.92\%} & 87.38\%\std{2.17\%} & 86.84\%\std{1.73\%} \\
& \textbf{Ours} &
\textbf{72.69\%\std{1.93\%}} & \textbf{78.03\%\std{2.47\%}} & \textbf{86.88\%\std{1.11\%}} &
\textbf{81.05\%\std{1.79\%}} & \textbf{76.12\%\std{0.95\%}} & \textbf{84.30\%\std{2.22\%}} &
\textbf{92.18\%\std{1.57\%}} & \textbf{87.52\%\std{2.14\%}} & \textbf{86.58\%\std{1.19\%}} \\

\bottomrule[1.5pt]
\end{tabular}
}
\caption{\footnotesize 
\textbf{Empirical comparison of prefix-level hallucination detection methods.}
We compare representative baselines~\cite{zhang2025icr,sriramanan2024llm} that perform hallucination detection using internal model representations.
Same as Table~\ref{tab: step}, global-mean aggregation baseline is included.
All methods are trained with identical prefix-level supervision for fair comparison.
We report results in two settings: \emph{Local}, which averages prefix-level predictions over all steps in the CoT, and \emph{Final}, which uses the prefix state at the final reasoning step to assess hallucination in the overall CoT outcome.}
\label{tab: prefix}
\vspace{-1em} %
\end{table*}


\subsection{Key Findings}
We next report results for prefix-level hallucination modeling.
Specifically, we present quantitative evaluations based on empirical data to assess model behavior, and complement them with qualitative case analyses to illustrate prefix-level dynamics.

\paragraph{Dynamic Evaluation of Prefix-level Behavior}

Table~\ref{tab: prefix} reports prefix-level hallucination detection results under the \emph{Local} and \emph{Final} settings.
Under \emph{Final} evaluation, our approach attains AUCs of $72.7\%$, $81.1\%$, and $92.2\%$ on \texttt{LLaMA}, \texttt{Qwen}, and \texttt{DeepSeek}, respectively, outperforming all baselines.
In contrast, several baselines show notable discrepancies between Local and Final performance, despite comparable aggregate scores.

\begin{figure}[h]
\vspace{-1em} %
\centering
\includegraphics[width=0.5\textwidth]{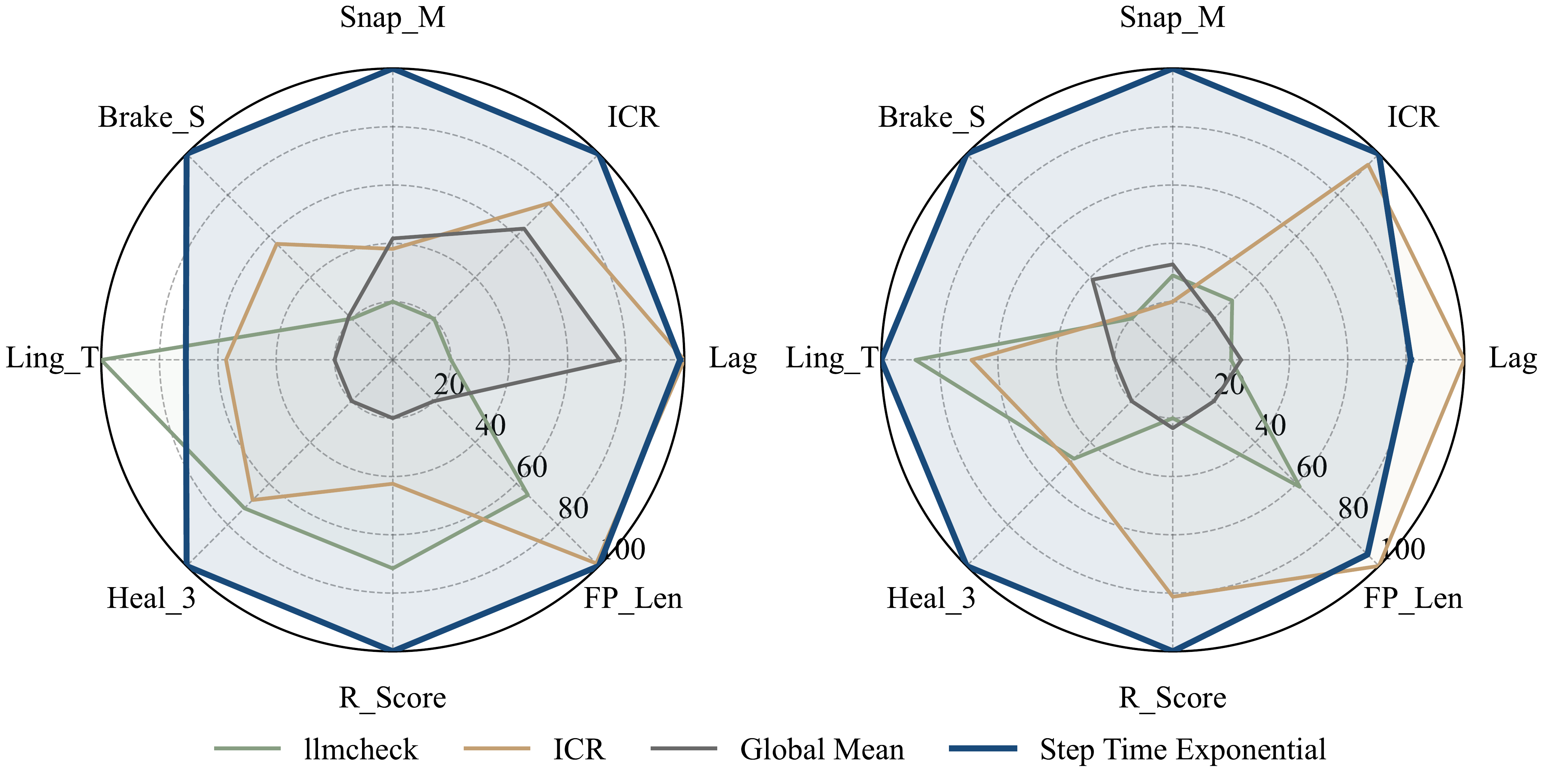}
\vspace{-2em} %
\caption{\footnotesize 
Radar visualization of eight dynamic metrics for prefix-level hallucination evaluation, all normalized to a $[0,100]$ scale.
Left: \texttt{LLaMA}; Right: \texttt{Qwen}.
Detailed definitions of all metrics are provided in Appendix~\ref{appx:dynamic_metrics}.
}
\label{fig: 8metric}
\vspace{-0.6em} %
\end{figure}

However, these metrics largely reflect average classification performance over prefixes.
They provide limited insight into how predictions evolve when hallucination states start or end, or how errors persist and recover along the trajectory.
To this end, we further evaluate models using a set of dynamic metrics that focus on hallucination onset, recovery, and false-positive structure.
As summarized in Figure~\ref{fig: 8metric}, our method consistently performs well across all eight dynamic indicators, reflecting coherent behavior throughout state transitions.

\obsbox{\textbf{Obs 4.} Incorporating step-conditioned prefix state estimation provides logical constraints that improve probe behavior beyond static prefix representations.}

Notably, although the global-mean baseline achieves relatively high AUC and appears numerically close to our method, its performance drops substantially under logic-oriented dynamic metrics.
This indicates that strong aggregate scores alone are insufficient to ensure coherent behavior during hallucination state transitions.
We attribute this gap to the lack of explicit dependency between step-level judgments and prefix-level state estimation during training,  in line with Property~IV.

We argue that a prefix-level hallucination state should be influenced by the current step judgment, since state transitions are induced by step-wise evidence.
At the same time, once a prefix state is formed, it naturally conditions the subsequent reasoning trajectory and thus affects the content and difficulty of later steps.
These two directions form a coupled feedback loop between step-level judgments and prefix-level states.

\obsbox{\textbf{Obs 5.}
Prefix states change in response to step-wise predictions and, once formed, systematically influence subsequent step-level behavior, revealing a bidirectional dependency along the reasoning trajectory.
}
\paragraph{Qualitative Analysis of Prefix-level Trajectories}

\begin{figure*}[h]
  \centering

  \begin{subfigure}[h]{0.48\textwidth}
        \centering
        \includegraphics[width=\textwidth,,height=0.5\textwidth]{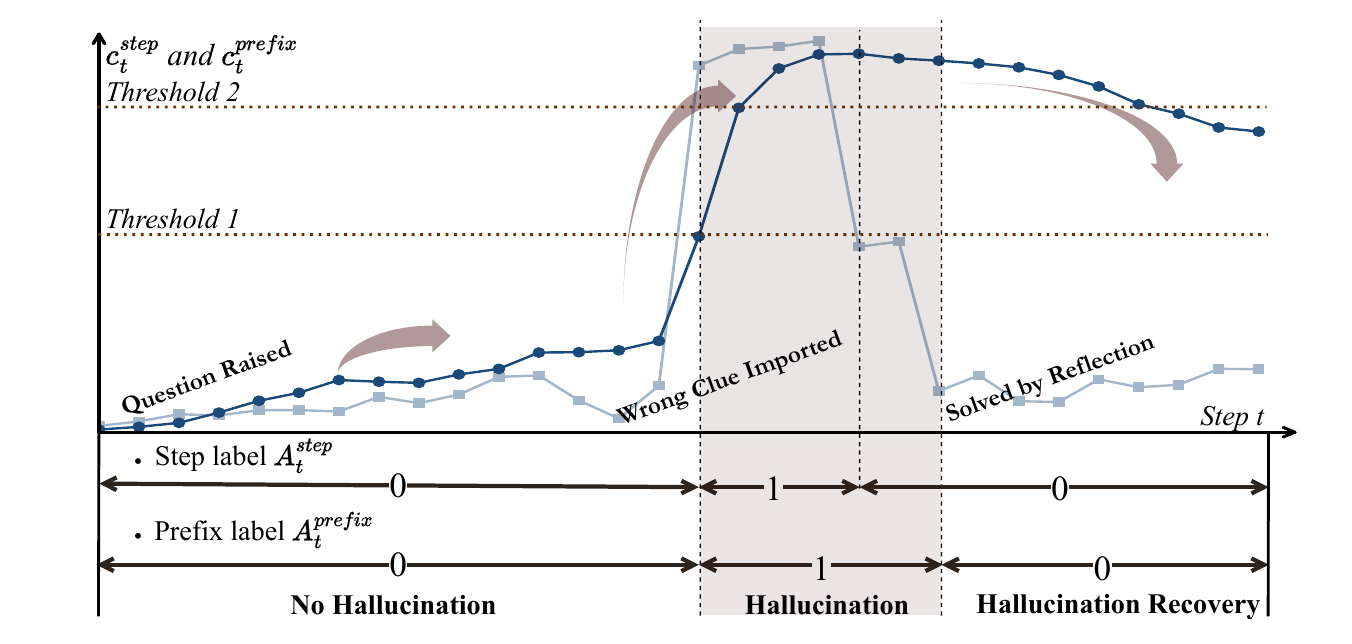}
        \caption{Error recovery after transient hallucination.}
        \label{fig:case4a}
    \end{subfigure}
    \begin{subfigure}[h]{0.48\textwidth}
        \centering
        \includegraphics[width=\textwidth,,height=0.5\textwidth]{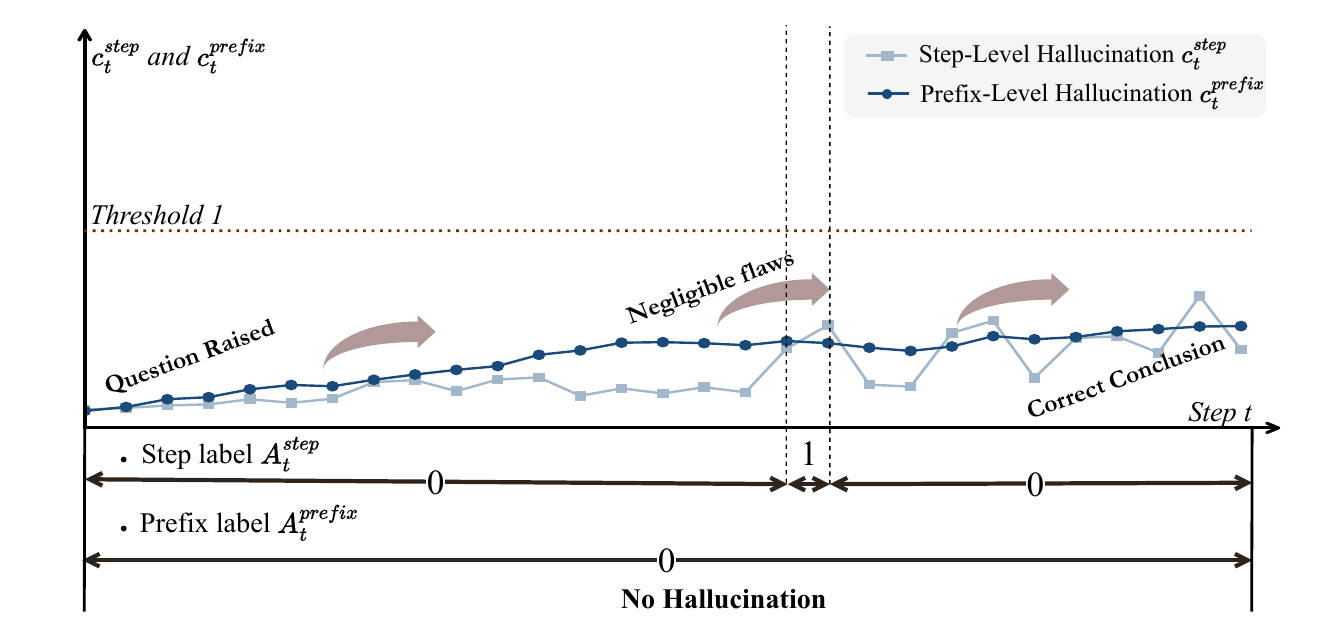}
        \caption{Consistently correct reasoning.}
        \label{fig:case4b}
    \end{subfigure}
    \begin{subfigure}[h]{0.48\textwidth}
        \centering
        \includegraphics[width=\textwidth,,height=0.5\textwidth]{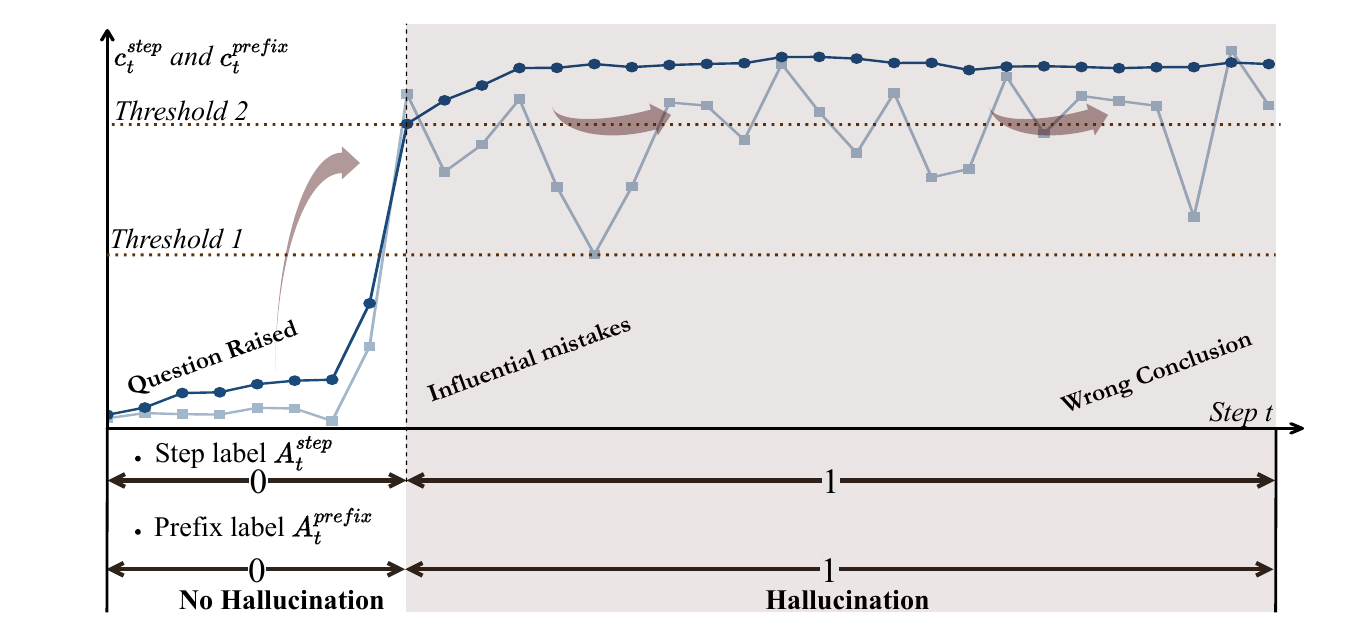}
        \caption{Case 3: Error occurs and persists.}
        \label{fig:case4c}
    \end{subfigure}
    \begin{subfigure}[h]{0.48\textwidth}
        \centering
        \includegraphics[width=\textwidth,,height=0.5\textwidth]{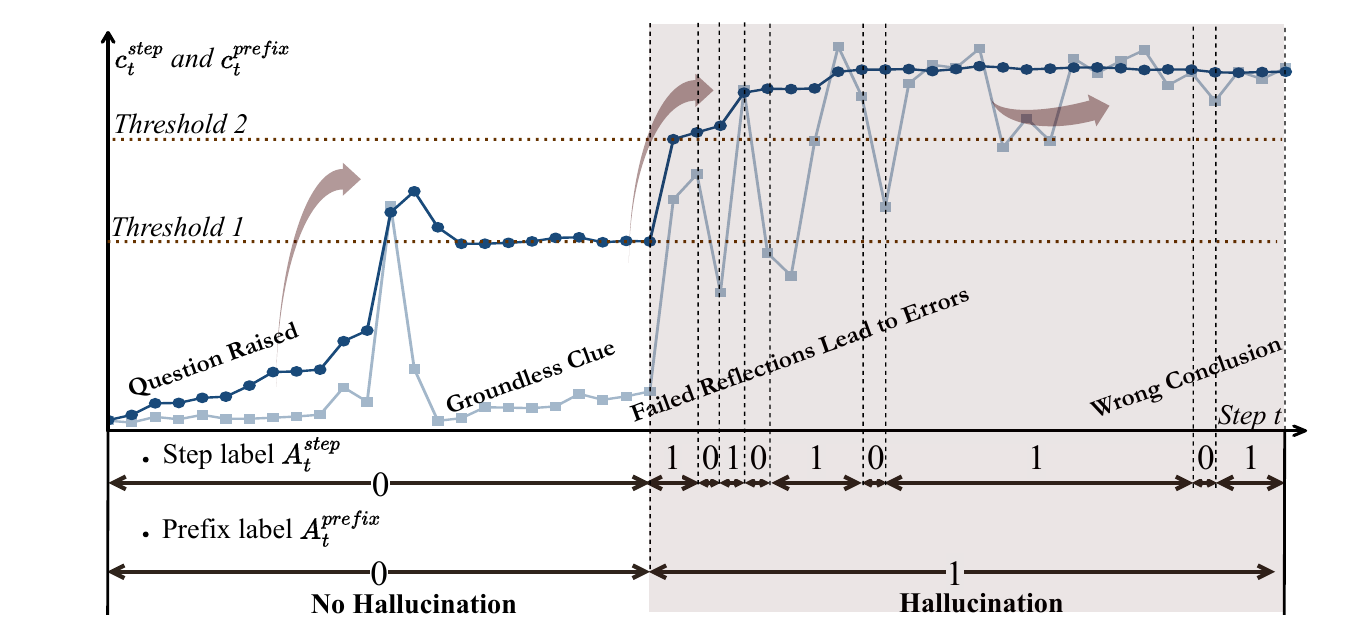}
        \caption{Case 4: Error occurs and correction fails.}
        \label{fig:case4d}
    \end{subfigure}
    \vspace{-0.5em} %
\caption{\footnotesize 
Qualitative examples of prefix-level hallucination behavior on real CoT trajectories.
Each subplot shows the evolution of step-level confidence ($c_t^{\mathrm{step}}$) and prefix-level confidence ($c_t^{\mathrm{prefix}}$) over reasoning steps.
The labels for each case are shown below the plots, and the red shaded regions indicate steps where the prefix state is labeled as hallucinated.
}
  \label{fig: case}
  \vspace{-1.5em}
\end{figure*}

Quantitative metrics provide a compact summary of performance, but they can obscure \emph{how} hallucination states evolve and interact with step-wise signals along a reasoning trajectory.
To complement our quantitative findings, we qualitatively examine representative CoT trajectories in Figure~\ref{fig: case}.
(Full trajectories with subsequent human annotations are provided in Appendix~\ref{appx: case}.)

The trajectories show a clear difference between how prefix-level states increase and how they decrease.
In Case~\ref{fig:case4a}, step-level confidence reacts quickly to an incorrect step and the following correction.
However, the prefix-level confidence decreases more slowly and only returns to a low level after several consecutive corrective steps.
In contrast, Case~\ref{fig:case4d} shows that although step-level confidence fluctuates and sometimes drops, the prefix-level confidence stays high for a long period and is difficult to reduce.
This suggests that once prefix-level hallucination builds up, it is not easily removed by short or isolated corrections.
\obsbox{\textbf{OBS 6.} Prefix-level hallucination shows asymmetric behavior: it can rise quickly after step-level errors, but decreases only after sustained corrective evidence.}

The trajectories also show that step-level confidence and prefix-level state do not always change together.
In Case~\ref{fig:case4b}, step-level confidence contains small spikes, but the prefix-level confidence remains low throughout the trajectory.
This indicates that isolated step-level anomalies do not necessarily cause a prefix-level hallucinated state.
On the other hand, in Case~\ref{fig:case4a}, step-level confidence drops soon after correction, while the prefix-level confidence remains high and decreases gradually.
These cases suggest that prefix-level confidence reflects accumulated evidence over multiple steps rather than the step-level signal at a single position.
\obsbox{\textbf{OBS 7.} Step-level confidence may change sharply without flipping the prefix state, while prefix-level recovery can lag behind step-level correction.}
\section{Related Work}
\textbf{Hallucination Detection and Mitigation.}
Prior work on hallucination in large language models mainly focuses on detection and mitigation at the output or training level. Detection methods rely on response consistency~\cite{farquhar2024detecting}, uncertainty estimation~\cite{shen2024thermometer}, or verifier-based factual checking~\cite{jiang2024graph}, while mitigation approaches include retrieval-augmented generation~\cite{jeong2024adaptive}, self-refinement~\cite{he2025self}, and alignment-based training such as RLHF and DPO~\cite{xu2025contrastive}. Although effective in reducing factual errors, these methods are typically retrospective or coarse-grained~\cite{cheng2025cot}, and do not explicitly model how hallucination evidence emerges, accumulates, or propagates during long chain-of-thought reasoning~\cite{gan2025rethinking}.

\textbf{Interpretability and Probing of Hallucination.}
Recent interpretability studies suggest that hallucination corresponds to systematic patterns in internal representations rather than random decoding noise~\cite{marks2024geometry}. Analyses based on hidden states, attention, and probing reveal that factuality, uncertainty, and reasoning reliability are often decodable from intermediate layers~\cite{bao2025probing}. However, most existing approaches analyze static representations or isolated reasoning steps~\cite{suresh2025cross}, treating probe predictions independently and overlooking the temporal dependency between local signals and the global reasoning state in long chain-of-thought settings~\cite{mao2025temporalizing}.
\vspace{-0.5em} %
\section{Conclusion}
\vspace{-0.5em} %
In this work, we argue that hallucination in long chain-of-thought reasoning is best understood as a temporally evolving latent state, rather than a collection of isolated local errors. This state-centric perspective shifts hallucination analysis from static detection to modeling the dynamics of reasoning itself, enabling more principled and interpretable assessment of long-form reasoning reliability.

\section{Limitations}

This work focuses on long-CoT reasoning settings where explicit intermediate steps are available. In cases where reasoning is implicit, compressed, or not externally exposed, the proposed step-level and prefix-level signals may be less directly observable.

Our approach relies on access to intermediate hidden states of the underlying language model and is therefore not directly applicable to black-box or API-only settings. Moreover, although we observe consistent trends across multiple model families, the optimal choice of probing layers and configurations may vary across architectures.

Finally, this work is primarily concerned with hallucination detection rather than mitigation. How prefix-level hallucination signals can be leveraged for active intervention, correction, or controlled regeneration during inference remains an open direction for future research.

\bibliography{custom}

\clearpage

\appendix
This appendix provides supplementary details to support the methodology and experimental results presented in the main text.
\begin{enumerate}
    \item \textbf{Appendix}~\ref{appx:data_validation} elaborates on the rigorous data validation mechanism used to ensure dataset quality.
    \item \textbf{Appendix}~\ref{appx:theoretical_analysis} provides a theoretical analysis of representation bias and the state-space formulation of hallucination.
    \item \textbf{Appendix}~\ref{appx:method_variants} describes the specific aggregation variants and feature strategies designed for the model.
    \item \textbf{Appendix}~\ref{appx:exp_analysis} presents comprehensive supplementary experimental analyses, specifically detailing the baseline methods (Appendix~\ref{appx:baseline_methods}) , the dynamic evaluation metrics (Appendix~\ref{appx:dynamic_metrics}) , the layer-wise performance analysis (Appendix~\ref{appx:layer_analysis}) , and the qualitative case studies (Appendix~\ref{appx: case}).
\end{enumerate}

\section{Data Validation Mechanism}
\label{appx:data_validation}

To ensure the high quality and logical self-consistency of the hallucination detection dataset, we implemented a strict \textbf{three-level validation mechanism} following the automated annotation process. This mechanism combines rule-based logical verification with expert manual review, aiming to eliminate samples containing logical paradoxes and semantic conflicts generated during annotation.

\subsection{Answer-Aware Semantic Consistency}
\label{appx:answer_aware}

Before proceeding with step-level annotation, we first established a global ``answer correctness'' benchmark. During the annotation phase, we enforced the model to follow an \textbf{Answer-Aware} evaluation paradigm.

Traditional string matching methods struggle to handle cases of semantic equivalence (such as paraphrasing or unit conversion) and can easily introduce noise. Therefore, we deployed Claude Sonnet 4.5 as an independent \emph{Semantic Judge}. By comparing the model's final generated answer ($A_{pred}$) with the standard answer ($A_{gold}$), the system outputs a global correctness label $Y \in \{Correct, Incorrect\}$.

Specifically, the process consists of two stages. First, for \emph{semantic equivalence arbitration}, both $A_{pred}$ and $A_{gold}$ are provided to Claude Sonnet 4.5, which determines whether the two answers are semantically equivalent based solely on meaning, and outputs the corresponding correctness label $Y$. This judgment follows several supplementary equivalence principles: format differences such as whitespace or punctuation are ignored; common variants of multiple-choice options (e.g., ``(A)'' vs.\ ``A'') are treated as equivalent; different numerical expressions (e.g., ``5.0'' vs.\ ``five'') are unified; and paraphrased phrases or sentences are recognized as equivalent. Through this design, the evaluation focuses on semantic content rather than superficial form.

Second, the correctness label $Y$ is injected as a prior condition into the subsequent annotation system prompt. Under this \textbf{Answer-Aware} strategy, the annotation model is forced to perform a form of reverse consistency reasoning. When $Y=Incorrect$, the model must identify the turning point in the reasoning chain that caused the deviation from factual correctness and ensure that the final cumulative state reflects this error. Conversely, when $Y=Correct$, even if the reasoning process exhibits temporary fluctuations, the model must verify whether the chain ultimately undergoes effective self-correction and returns to a correct trajectory.

This mechanism reduces semantic contradictions at the source, preventing cases where a correct final result is labeled as a full hallucination, or where an incorrect result is mistakenly treated as fully correct.

\subsection{Logical Consistency Check}
\label{appx:logical_check}

Hallucination annotations in large language models often suffer from \emph{local--global incoherence}. To eliminate this issue, we formalized the annotation rules as a set of state transition constraints.

Let the reasoning sequence length be $T$. At step $t$, the local hallucination label is denoted as $s_t \in \{0,1\}$, where $1$ indicates the presence of a hallucination, while the cumulative hallucination label is $c_t \in \{0,1\}$, where $1$ indicates that the reasoning path has been contaminated. Any sample violating the constraints defined below is considered logically invalid and is directly \textbf{discarded}.

\subsubsection*{Rule 1: Terminal Consistency Constraint}

The most basic requirement concerns the final state of the reasoning chain. According to our verification logic, the cumulative hallucination state at the terminal step ($c_T$) must be mutually exclusive with the global answer correctness label $Y$, namely:
\begin{equation}
c_T = 1 \iff Y = \text{Incorrect},
\end{equation}
\begin{equation}
c_T = 0 \iff Y = \text{Correct}.
\end{equation}

The verification principle is straightforward. If the model's final answer is correct ($Y=Correct$), the reasoning chain must end on a valid path, implying $c_T=0$. Conversely, if the final answer is incorrect, the erroneous reasoning must have persisted to the end, and thus $c_T=1$. Any sample exhibiting inconsistency between these two signals—such as a correct answer marked as a final hallucination, or an incorrect answer marked as fully correct—is regarded as a severe logical conflict and is removed from the dataset.

\subsubsection*{Rule 2: Transition Validity Detection}

Beyond the terminal condition, we further scrutinize transitions between cumulative reasoning states, namely $c_{t-1}=1 \rightarrow c_t=0$ and $c_{t-1}=0 \rightarrow c_t=1$. According to our annotation definition, such transitions must be grounded in the logical properties of the current step.

We distinguish four possible transition modes:
\begin{enumerate}
    \item \textbf{Valid Recovery}: $(c_{t-1}=1 \land c_t=0) \land s_t=0$.  
    In this case, the model was previously in a hallucinated state, but produces a correct factual statement or valid deduction at step $t$, thereby restoring the reasoning path. This transition is logically sound.

    \item \textbf{Anomalous Recovery}: $(c_{t-1}=1 \land c_t=0) \land s_t=1$.  
    Here, the model remains factually or logically incorrect at step $t$, yet the cumulative state is judged as having recovered. This constitutes a logical paradox in which multiple errors allegedly lead to correctness.

    \item \textbf{Valid Degradation}: $(c_{t-1}=0 \land c_t=1) \land s_t=1$.  
    In this scenario, the reasoning path was previously correct, but the current step introduces a factual or logical error, causing a consistent degradation.

    \item \textbf{Spurious Degradation}: $(c_{t-1}=0 \land c_t=1) \land s_t=0$.  
    The cumulative state is marked as entering hallucination despite the absence of any error at the current step, indicating a lack of causal justification.
\end{enumerate}

Based on these definitions, we implement a strict cleaning strategy that targets two severe failure patterns. A \emph{Severe Epiphany} occurs when an Anomalous Recovery transition is detected and the reasoning chain has remained in a cumulative hallucinated state for $n$ consecutive steps (typically $n \ge 5$) prior to the transition, implying a miraculous return to correctness without any valid corrective reasoning. A \emph{Severe Degradation} occurs when a Spurious Degradation transition follows a long stretch of cumulative correctness (again typically $n \ge 5$), indicating an unjustified collapse of the reasoning state. Any sample exhibiting either pattern is considered to have unreliable annotations and is directly removed from the dataset.

\subsection{Manual Verification}
\label{appx:manual_verification}

After automated logical filtering, we further conduct manual verification through stratified sampling to assess fine-grained annotation quality. Specifically, we randomly sample 5\% of the logically self-consistent dataset for expert review.

The review focuses on two aspects. First, experts examine potential false positives or false negatives in the local hallucination labels $s_t$, with particular attention to steps involving numerical computation or complex logical inference. Second, for segments labeled as recovery (either valid or anomalous), experts verify whether the step truly exhibits explicit or implicit correction semantics—such as acknowledging a mistake or revising earlier assumptions—rather than coincidentally guessing the correct answer.

Quantitative analysis shows a high level of agreement between automated annotation and human judgment, with an observed consistency rate exceeding 96\% on the sampled data. This result indicates that Claude Sonnet 4.5, when operating under the proposed \textbf{Answer-Aware} prompting framework and strict logical constraints, achieves human-level reliability in reasoning evaluation.

\section{Theoretical Analysis: Representation Bias and Hallucination Dynamics}
\label{appx:theoretical_analysis}

This appendix provides a theoretical analysis of the structural biases in standard representation methods and formally defines the properties required for modeling prefix-level hallucination. We first identify the limitations in common aggregation schemes (\textbf{Properties I and II}) and then introduce a state-space formulation that motivates our design choices (\textbf{Properties III and IV}).

\subsection{Step- and Token-level Representation Bias in Long CoT Reasoning}
\label{appx:rep_bias}

In autoregressive language models, reasoning is performed at the token level rather than at higher-level semantic units such as reasoning steps.
Formally, given an input $x$, a long chain-of-thought (CoT) reasoning process is represented as a sequence of reasoning steps
\begin{equation}
C(x) = (s_1, s_2, \ldots, s_T),
\end{equation}
where each step $s_t \in \Sigma^*$ is a contiguous segment of generated text.
Each reasoning step $s_t$ consists of a sequence of tokens
\begin{equation}
s_t = (w_{t,1}, w_{t,2}, \ldots, w_{t,L_t}),
\end{equation}
where $L_t$ denotes the number of tokens in step $s_t$.

During generation, the language model produces a hidden representation for each token at every layer.
Let $\mathbf{h}^{(l)}_{t,j} \in \mathbb{R}^d$ denote the hidden state of the $j$-th token in step $s_t$, taken from the $l$-th layer of the model, where $d$ is the hidden dimension.
Thus, the hidden states associated with step $s_t$ at layer $l$ form a matrix
\begin{equation}
\label{eq:H_full_sequence}
\mathbf{H}^{(l)}_t
=
\big(
\mathbf{h}^{(l)}_{t,1}, \mathbf{h}^{(l)}_{t,2}, \ldots, \mathbf{h}^{(l)}_{t,L_t}
\big)
\in \mathbb{R}^{L_t \times d}.
\end{equation}

After generating the full reasoning trajectory, the complete hidden representation at layer $l$ consists of all token-level hidden states concatenated in generation order:
\begin{equation}
\mathbf{H}^{(l)}
=
\big(
\mathbf{H}^{(l)}_1,
\mathbf{H}^{(l)}_2,
\ldots,
\mathbf{H}^{(l)}_T
\big)
\in
\mathbb{R}^{(\sum_{t=1}^{T} L_t) \times d}.
\end{equation}
Importantly, the model itself does not maintain step-level hidden states; any step-level representation is obtained by aggregating token-level representations from $\mathbf{H}^{(l)}$.

In practice, existing approaches typically construct representations for probing or analysis by aggregating token hidden states.
Below, we present two observations showing that commonly used aggregation schemes introduce structural biases that attenuate information newly introduced at the current reasoning step.

\paragraph{Property I: Cross-step aggregation attenuates signals from later reasoning steps.}
A common strategy to represent the reasoning state after step $s_t$ is to average the hidden states of all tokens generated so far:
\begin{equation}
\mathbf{z}_t
=
\frac{1}{\sum_{i=1}^{t} L_i}
\sum_{i=1}^{t}
\sum_{j=1}^{L_i}
\mathbf{h}^{(l)}_{i,j},
\qquad
\mathbf{z}_t \in \mathbb{R}^d.
\end{equation}
Under this construction, the total contribution of step $s_t$ to $\mathbf{z}_t$ is proportional to $\frac{L_t}{\sum_{i=1}^{t} L_i}$. As the reasoning trajectory grows longer, the total number of prefix tokens $\sum_{i=1}^{t} L_i$ typically becomes much larger than the number of tokens in the current step $L_t$.
Consequently, even if all tokens in step $s_t$ introduce systematic semantic changes or hallucinated content, their combined influence on the aggregated representation $\mathbf{z}_t$ is diluted by the large number of earlier tokens.

This attenuation arises from the shape mismatch between the underlying representation $\mathbf{H}^{(l)} \in \mathbb{R}^{(\sum_i L_i)\times d}$ and its pooled proxy $\mathbf{z}_t \in \mathbb{R}^d$, where newly added rows corresponding to the current step contribute only a small fraction to the final vector.

\paragraph{Property II: Within-step averaging downweights information from later tokens.}
A similar bias appears when constructing a step-level representation by aggregating token hidden states within a single reasoning step.
Consider the hidden states $\mathbf{H}^{(l)}_t$ that form the components of the full sequence representation $\mathbf{H}^{(l)}$ defined in~\eqref{eq:H_full_sequence}, which correspond to step $s_t$ at layer $l$.

Due to the autoregressive nature of generation, later token hidden states implicitly encode information from earlier tokens.
At an abstract level, this dependence can be expressed as
\begin{equation}
\mathbf{h}^{(l)}_{t,j}
\;\approx\;
\mathbf{h}^{(l)}_{t,0}
+
\sum_{k=1}^{j}
\mathbf{u}^{(l)}_{t,k},
\qquad
\mathbf{u}^{(l)}_{t,k} \in \mathbb{R}^d,
\end{equation}
where $\mathbf{u}^{(l)}_{t,k}$ represents the incremental contribution introduced when generating the $k$-th token of step $s_t$. Since $\mathbf{h}^{(l)}_{t,0}$ contributes equally to all token positions, it does not affect the relative weighting induced by uniform averaging and is omitted in the following derivation.

Due to the autoregressive nature of language models, information flow within a
reasoning step is inherently asymmetric across token positions.
Earlier tokens are generated without access to later tokens, whereas later
tokens are conditioned on and can integrate information from all preceding
tokens in the same step.

As a result, token representations toward the end of a reasoning step tend to
encode a more complete summary of the step-level semantics, including potential
inconsistencies or hallucinated content introduced during the step.
This asymmetry suggests that later tokens may carry more diagnostic information
for assessing the reliability of the current reasoning step, rendering uniform
averaging across token positions a potentially suboptimal aggregation strategy.

If a step-level representation is obtained by uniformly averaging token hidden states,
\begin{equation}
\tilde{\mathbf{h}}^{(l)}_t
=
\frac{1}{L_t}
\sum_{j=1}^{L_t}
\mathbf{h}^{(l)}_{t,j},
\qquad
\tilde{\mathbf{h}}^{(l)}_t \in \mathbb{R}^d,
\end{equation}
then substituting the autoregressive form yields
\begin{equation}
\tilde{\mathbf{h}}^{(l)}_t
=
\sum_{k=1}^{L_t}
\Big(1 - \frac{k-1}{L_t}\Big)
\mathbf{u}^{(l)}_{t,k}.
\end{equation}
This expression makes explicit that information introduced at later token positions is assigned smaller weights.
As $L_t$ increases, the resulting step-level vector $\tilde{\mathbf{h}}^{(l)}_t$ becomes increasingly dominated by contributions from early tokens in the step, reducing sensitivity to semantic shifts or errors that arise near the end of the step.

Taken together, these observations highlight a structural limitation in common aggregation schemes.
Although the underlying hidden representation $\mathbf{H}^{(l)} \in \mathbb{R}^{(\sum_t L_t)\times d}$ contains fine-grained, token-level information, mapping it to a step-level vector in $\mathbb{R}^d$ via uniform averaging introduces a strong bias toward earlier context.
This bias can obscure newly introduced information in the current reasoning step, making step-level hallucination signals difficult to detect in long CoT reasoning.

\subsection{A State-Space View of Prefix-level Hallucination}
\label{appx:prefix_properties_math}
In this section, we provide a mathematical discussion of the properties required for prefix-level hallucination modeling.
Our analysis adopts a state-space perspective: hallucination is treated as a latent reasoning state, while step-level signals are regarded as local observations.
Importantly, this discussion is independent of any specific model architecture or loss design, and instead focuses on structural constraints implied by this formulation.
Let  $\{\mathbf{h}_t\}_{t=1}^T$ denote the sequence of hidden representations produced along a chain-of-thought (CoT) reasoning trajectory.
As discussed in the main text, transitions between consecutive reasoning steps can be written in an incremental form:
\begin{equation}
\mathbf{h}_{t+1} = \mathbf{h}_t + \mathbf{u}_{t+1},
\end{equation}
where $\mathbf{u}_{t+1}$ represents the state change introduced by step $s_{t+1}$.

We model prefix-level hallucination using a latent binary variable $Z_t \in {0,1}$, which indicates whether the reasoning prefix $s_{1:t}$ has entered a hallucinated state.
The prefix-level hallucination score $c_t^{\mathrm{prefix}}$ can therefore be interpreted as an estimate of
\begin{equation}
c_t^{\mathrm{prefix}} \approx \mathbb{P}(Z_t = 1 \mid \mathcal{F}_t),
\end{equation}
where $\mathcal{F}_t$ denotes all information available up to step $t$.

In contrast, step-level hallucination indicators $H_{t+1}^{\mathrm{step}}$ and their corresponding scores $c_{t+1}^{\mathrm{step}}$ are associated with the newly introduced increment $\mathbf{u}_{t+1}$.
From this perspective, $c_{t+1}^{\mathrm{step}}$ serves as a local and potentially noisy observation of whether the current update deviates from valid reasoning behavior, rather than a direct estimate of the global reasoning state.

This distinction naturally leads to a state–observation separation: prefix-level hallucination reflects a latent state of the reasoning process, whereas step-level hallucination provides local evidence about individual updates.

\paragraph{Property III: Temporal Coherence}
The first property concerns temporal continuity.
Since $Z_t$ represents a latent reasoning state rather than an instantaneous event, it is reasonable to assume that its evolution is governed by a transition process with limited volatility.
In other words, the probability of switching between hallucinated and non-hallucinated states within a single reasoning step is typically small.

Recall that the prefix-level hallucination score
$c_t^{\mathrm{prefix}}$ is intended to approximate
$\mathbb{P}(Z_t = 1 \mid \mathcal{F}_t)$. Temporal continuity implies that, for the majority of steps, the expected change is bounded:
\begin{equation}
\mathbb{E}\!\left[
\left| c_{t+1}^{\mathrm{prefix}} - c_t^{\mathrm{prefix}} \right|
\mid \mathcal{F}_t
\right] \leq \epsilon .
\end{equation}
for some small constant $\epsilon > 0$.

This condition does not enforce monotonicity, nor does it forbid occasional larger updates.
Instead, it constrains the expected magnitude of state changes.
Such a constraint is consistent with the incremental nature of hidden-state transitions and with the interpretation of hallucination as a persistent reasoning condition rather than a rapidly fluctuating signal.

Allowing frequent large oscillations in $p_t$ would imply unrealistically high transition probabilities between latent states, which would undermine the notion of hallucination as a coherent reasoning state.

\paragraph{Property IV: Directional Consistency with Local Evidence}
Temporal continuity alone does not specify how the latent state should respond to new observations.
The second property therefore concerns directional consistency.

Prefix-level hallucination modeling should remain responsive to informative step-level evidence, allowing the hallucination score to both increase and decrease over time.
This flexibility is particularly important in long CoT reasoning, where later steps may revise or correct earlier incorrect assumptions.
As a result, the presence of a hallucinated step does not necessarily imply an incorrect final answer.

Formally, directional consistency requires that updates to the prefix-level hallucination score be statistically aligned with step-level evidence when such evidence is strong.
This can be expressed using a covariance constraint:
\begin{equation}
\mathrm{Cov}\left(
    c_{t+1}^{\mathrm{prefix}} - c_t^{\mathrm{prefix}}, \,
    c_{t+1}^{\mathrm{step}}
    \;\middle|\;
    \mathcal{F}_t
\right) \ge 0
\end{equation}

This condition ensures that strong step-level evidence for hallucination tends to increase the prefix-level score, while weak or absent evidence does not systematically drive it upward.
Importantly, this requirement does not impose a deterministic or monotonic relationship between step-level and prefix-level scores.
Negative updates remain admissible, reflecting the possibility of later self-correction.

\section{Method and Method Variants}
\label{appx:method_variants}

We design several aggregation variants to summarize token-level hidden states $\mathbf{h}$ during the reasoning process. These variants differ in the temporal scope they consider (current step $t$ vs.\ global history $1 \dots t$) and the weighting strategy used for aggregation.

\subsection{Step-level Aggregation}
\label{appx:step_aggregation}

\paragraph{Step Mean }
\begin{equation}
\mathbf{z}_t^{\mathrm{step\text{-}mean}}
=
\frac{1}{L_t}
\sum_{i=1}^{L_t}
\mathbf{h}_{t,i},
\end{equation}
This variant focuses only on the tokens generated within the current reasoning step $t$.
It computes the arithmetic mean $\mathbf{z}_t^{\mathrm{step\text{-}mean}}$ of the hidden-state vectors $\mathbf{h}_{t,i}$ for all $L_t$ tokens in the current step.

The resulting representation reflects the average semantic content of the current reasoning step.
It assumes that all tokens $i$ within the step contribute equally to judging the correctness of this step.

\paragraph{Step Time Exp}
\begin{equation}
w_{t,i} = \frac{i-1}{L_t-1},
\qquad
\alpha_{t,i}
=
\frac{\exp(w_{t,i})}
{\sum_{j=1}^{L_t} \exp(w_{t,j})},
\end{equation}
\begin{equation}
\mathbf{z}_t^{\mathrm{step\text{-}exp}}
=
\sum_{i=1}^{L_t}
\alpha_{t,i}\,\mathbf{h}_{t,i}.
\end{equation}

This variant considers only the current reasoning step $t$ and aggregates token representations $\mathbf{h}_{t,i}$ using exponentially increasing weights $\alpha_{t,i}$.
Tokens closer to the end of the step (where the relative weight $w_{t,i} \approx 1$) receive higher weights.

The resulting representation $\mathbf{z}_t^{\mathrm{step\text{-}exp}}$ emphasizes the later part of the current step.
It reflects the intuition that tokens generated near the end of a reasoning step have a more complete view of the entire step’s information, making them more informative for judging whether the step contains hallucinations.

\subsection{Global Aggregation}
\label{appx:global_aggregation}

\paragraph{Global Mean }
\begin{equation}
\mathbf{z}_t^{\mathrm{global\text{-}mean}}
=
\frac{1}{n_t}
\sum_{k=1}^{t}
\sum_{i=1}^{L_k}
\mathbf{h}_{k,i}.
\end{equation}
This variant aggregates all token hidden states $\mathbf{h}_{k,i}$ from the beginning of the reasoning process ($k=1$) up to the current time $t$.
It computes a simple average $\mathbf{z}_t^{\mathrm{global\text{-}mean}}$ over all $n_t$ past tokens.

The resulting vector captures the average global context of the entire reasoning history.
It treats early and recent information as equally important.

\paragraph{Global Linear }
\begin{equation}
\mathbf{z}_t^{\mathrm{global\text{-}lin}}
=
\frac{
\sum_{k=1}^{t}
\sum_{i=1}^{L_k}
\left(
\sum_{m=1}^{k-1} L_m + i
\right)
\mathbf{h}_{k,i}
}{
\sum_{k=1}^{t}
\sum_{i=1}^{L_k}
\left(
\sum_{m=1}^{k-1} L_m + i
\right)
}.
\end{equation}
This variant also considers all tokens up to the current time $t$, but assigns linearly increasing weights based on temporal proximity.
Specifically, the weight for $\mathbf{h}_{k,i}$ is proportional to its global token index $(\sum L_m + i)$.
More recent tokens receive higher weights, while earlier tokens receive lower weights.

This acts as a mild forgetting mechanism.
It assumes that recent reasoning context is more relevant than distant history, while still retaining information from earlier steps.

\paragraph{Global Exp }
\begin{equation}
\omega_{k,i}
=
\exp\!\left(
\gamma \left(
\sum_{m=1}^{k-1} L_m + i
\right)
\right),
\end{equation}
\begin{equation}
\mathbf{z}_t^{\mathrm{global\text{-}exp}}
=
\frac{
\sum_{k=1}^{t}
\sum_{i=1}^{L_k}
\omega_{k,i}\,\mathbf{h}_{k,i}
}{
\sum_{k=1}^{t}
\sum_{i=1}^{L_k}
\omega_{k,i}
},
\qquad \gamma = 0.003.
\end{equation}
This variant aggregates all past token hidden states using exponentially increasing weights $\omega_{k,i}$.
Tokens closer to the current time dominate the aggregation $\mathbf{z}_t^{\mathrm{global\text{-}exp}}$, with the decay rate controlled by $\gamma$.

This represents a strong focus on recent information.
It rapidly downweights distant history and assumes that the correctness of the current reasoning state mainly depends on the most recent reasoning process.

\subsection{Alternative Step-level Representation Methods}
\label{appx:appendix_alternative_features}

In addition to the time-aware exponential weighting method proposed in Section~\ref{sec: step-level}, we investigated a diverse set of feature aggregation strategies to construct the step-level representation $\mathbf{z}_t$. These methods aim to capture different aspects of the reasoning process, such as worst-case uncertainty or specific token saliency, by manipulating the set of token-level hidden states $\{\mathbf{h}_{t,1}^{(l)}, \dots, \mathbf{h}_{t,L_t}^{(l)}\}$ within a reasoning step $s_t$. We categorize these approaches into statistical pooling, uncertainty-aware aggregation, and scalar distribution features.

\subsubsection{Statistical Pooling Strategies}

\paragraph{Max Pooling.}
Unlike average pooling, which may dilute strong signals with neutral tokens, max pooling extracts the most salient feature activation across the step. We compute the element-wise maximum of the hidden states:
\begin{equation}
    \mathbf{z}_t^{\text{max}} = \max_{j=1}^{L_t} \left\{ \mathbf{h}_{t,j}^{(l)} \right\} \in \mathbb{R}^d,
\end{equation}
where the maximum operation is applied independently to each dimension of the hidden vector.

\textit{Rationale:} This approach assumes that the presence of specific semantic features (e.g., a high activation in a "negation" or "conflict" neuron) is more diagnostic of a hallucinated state than the average context.

\paragraph{Last Token Selection.}
Given the autoregressive nature of LLMs, the final token of a step theoretically attends to and aggregates information from all preceding tokens in that step. We define the representation simply as:
\begin{equation}
    \mathbf{z}_t^{\text{last}} = \mathbf{h}_{t, L_t}^{(l)}.
\end{equation}

\textit{Rationale:} This serves as a baseline reflecting the model's immediate state before transitioning to the next reasoning step, positing that the final hidden state implicitly summarizes the local reasoning trajectory.

\subsubsection{Uncertainty-Aware Aggregation}
These methods leverage the model's output probabilities to weigh hidden states, prioritizing tokens where the model exhibits lower confidence. Let $p_{t,j}$ denote the probability of the $j$-th token in step $s_t$, derived from the logits.

\paragraph{Surprisal-Weighted Aggregation.}
We interpret the negative log-probability as surprisal, $S_{t,j} = -\log p_{t,j}$. To emphasize unexpected tokens, we compute a weighted sum of hidden states where weights are derived from the softmax of surprisals:
\begin{equation}
    \mathbf{z}_t^{\text{surp}} = \sum_{j=1}^{L_t} \alpha_{t,j} \mathbf{h}_{t,j}^{(l)}, \quad \text{where } \alpha_{t,j} = \frac{\exp(S_{t,j})}{\sum_{k=1}^{L_t} \exp(S_{t,k})}.
\end{equation}

\textit{Rationale:} Tokens with high surprisal often indicate points where the model deviates from robust reasoning or forces a low-probability generation. This aggregation focuses the representation on these high-risk pivot points.

\paragraph{Minimum Probability State.}
This method isolates the "weakest link" in the reasoning step. We select the hidden state corresponding to the token with the lowest assigned probability:
\begin{equation}
    \mathbf{z}_t^{\text{min}} = \mathbf{h}_{t, j^*}^{(l)}, \quad \text{where } j^* = \operatorname*{argmin}_{j \in \{1, \dots, L_t\}} p_{t,j}.
\end{equation}

\textit{Rationale:} In many cases, a single hallucinated entity or incorrect logical connector is accompanied by a sharp drop in local confidence. Using the minimum probability state prevents this signal from being washed out by high-confidence function words.

\paragraph{Bottom-5 Weighted Aggregation.}
To balance the focus on low-confidence tokens while mitigating noise from a single outlier, we consider the set of indices $\mathcal{J}_{\text{bot}}$ corresponding to the $k=5$ tokens with the lowest probabilities in the step (or all tokens if $L_t < 5$). We re-normalize their surprisal weights restricted to this subset:
\begin{equation}
    \mathbf{z}_t^{\text{bot5}} = \sum_{j \in \mathcal{J}_{\text{bot}}} \tilde{\alpha}_{t,j} \mathbf{h}_{t,j}^{(l)},
\end{equation}
where $\tilde{\alpha}_{t,j}$ is the softmax of surprisals computed only over $j \in \mathcal{J}_{\text{bot}}$.

\textit{Rationale:} This approach acts as a robust version of the minimum probability selector, capturing a cluster of uncertainty that often characterizes complex fabrications.

\subsubsection{Scalar Distribution Features}
In addition to high-dimensional hidden representations, we extract a low-dimensional vector $\mathbf{v}_t \in \mathbb{R}^{32}$ summarizing the statistical distribution of token probabilities within the step.

\paragraph{Feature Construction.}
We do not use hidden states for this set. Instead, we compute descriptive statistics of the sequence of token confidences $\{p_{t,1}, \dots, p_{t,L_t}\}$. These statistics include:
\begin{itemize}
    \item \textbf{Central Tendency and Dispersion:} Mean, median, standard deviation, and the ratio of standard deviation to the mean.
    \item \textbf{Extremes and Quantiles:} Minimum, maximum, range, and specific percentile values (e.g., 25th, 75th, 90th percentiles).
    \item \textbf{Threshold Ratios:} The proportion of tokens with confidence exceeding high thresholds (0.5, 0.7) or falling below low thresholds (0.3).
    \item \textbf{Dynamics:} First and last token confidences, the average confidence of the final third of the step, and the maximum distinct jump/drop between consecutive tokens.
\end{itemize}
Finally, we append a normalized 10-bin histogram of the confidence distribution.

\textit{Rationale:} These scalar features abstract away from semantic content to focus purely on the model's "metacognitive" signaling. Patterns such as high variance or a consistent downward trend in confidence often correlate with the onset of hallucination, providing a lightweight diagnostic signal.

\begin{table}[t]
\centering
\small
\setlength{\tabcolsep}{6pt}
\begin{tabular}{l ccc}
\toprule
\multirow{2}{*}{\textbf{Method}} & \multicolumn{3}{c}{\textbf{Llama-3.1-8B-Instruct}} \\
\cmidrule(lr){2-4}
 & \textbf{AUC} & \textbf{ACC} & \textbf{F1} \\
\midrule
Max Pooling & 86.23\% & 78.83\% & 70.50\% \\
Last Token & 85.54\% & 78.19\% & 69.17\% \\
Surprisal-Weighted & 86.62\% & 78.95\% & 70.49\% \\
Min Prob State & 83.69\% & 76.51\%& 67.15\% \\
Bottom-5 Weighted & 85.99\% & 78.57\% & 69.85\% \\
Scalar Features & 74.68\% & 71.81\% & 56.47\% \\
\bottomrule
\end{tabular}
\caption{Empirical comparison of alternative step-level representation strategies on Llama-3.1-8B-Instruct. All methods are evaluated under the same experimental settings as the main step-level probing experiments.}
\label{tab:appendix_alternative_results}
\end{table}


\section{Supplementary Experimental Analysis}
\label{appx:exp_analysis}
This section provides additional details on the experimental setup and results. We first describe the baseline methods and the dynamic evaluation metrics. Subsequently, we present a detailed layer-wise performance analysis to justify our feature selection strategy, followed by qualitative case studies.

\subsection{Baseline Methods}
\label{appx:baseline_methods}

This appendix summarizes representative white-box methods for truthfulness and hallucination detection in Large Language Models (LLMs). We focus on their core modeling assumptions and internal-state-based detection mechanisms.

\subsubsection{TTPD}

TTPD\cite{burger2024truth} reveals a universal, linear \emph{truthfulness direction} within Large Language Models (LLMs),  suggesting that the concept of truth is encoded in a structured and cross-domain manner within the model’s hidden states, independent of specific subject matter.
Based on this observation, the method introduces the \textbf{Truth and Polarity Direction training} algorithm to extract pure semantic signals from entangled internal representations.

Concretely, the approach identifies a two-dimensional plane in the activation space spanned by a \emph{truth} axis and a \emph{polarity} axis.
Within this plane, affirmative/negative and true/false statements form a clear rectangular distribution.
By constructing pairwise activation differences and applying the orthogonalization process, TTPD disentangles these intertwined semantic dimensions.
This geometric formulation enables robust discrimination between logically distinct statements, such as \emph{``Paris is in Germany''} (Affirmative False) and \emph{``Paris is not in Germany''} (Negative True).

\subsubsection{SAPLMA}

The core contribution of SAPLMA\cite{azaria2023internal} posits that LLMs often possess a form of \emph{self-awareness}, in the sense that even when generating hallucinated or incorrect outputs, their hidden states typically retain information about answer correctness.
This finding challenges the assumption that hallucinations arise purely from ignorance and provides theoretical support for detecting factual errors via internal activations.

To operationalize this insight, the authors introduce the \textbf{Self-Awareness Probe for Large Model Activations (SAPLMA)}. The method feeds the model with a set of true/false questions and extracts hidden-layer activations during response generation. These activations are averaged across all tokens, and a multi-layer perceptron (MLP) classifier is trained to map the resulting representations to a probability of truthfulness. Empirical results show that middle-to-late layers contain the most informative truth-related features, positioning SAPLMA as a representative white-box hallucination detection approach that directly leverages internal states rather than output probabilities.

\subsubsection{ICR Probe}

The ICR\cite{zhang2025icr} Probe adopts a dynamic perspective on hallucination detection, viewing hallucinations as an evolving process rather than a static property of a single activation.Instead of extracting a single hidden vector, the method tracks the trajectory of hidden states throughout the reasoning or generation process.

By measuring cosine similarity or representational drift between adjacent reasoning steps, the ICR Probe evaluates internal logical consistency over time.Sudden fluctuations or unstable transitions in hidden states are interpreted as signals that the model has begun deviating from factual knowledge.To further enhance robustness, the approach incorporates contrastive learning to distinguish high-consistency and low-consistency patterns, enabling more reliable detection in long-form generation scenarios.

\subsubsection{LLM-Check}

LLM-Check\cite{sriramanan2024llm} aims to reduce the computational cost and access constraints associated with existing hallucination detection methods by reframing detection as a geometric and spectral analysis problem.
Drawing on techniques from statistical physics and signal processing, the method applies eigen-analysis to quantify structural properties of hidden layer activations and attention maps.

Specifically, LLM-Check computes the Mean Log-Determinant of these internal representations to estimate the ``volume'' or degree of order in the latent space.
The study finds that truthful generations tend to exhibit compact and well-structured internal representations, whereas hallucinations correspond to expanded, disordered, or high-entropy states.
This spectral characterization enables hallucination detection within a single forward pass, achieving significant speedups by avoiding multi-sample decoding.
Moreover, through a proxy-model strategy, LLM-Check extends internal feature-based detection to black-box APIs such as GPT-4.

\subsection{Dynamic Evaluation Metrics for Prefix-level Hallucination}
\label{appx:dynamic_metrics}

Standard aggregate metrics (e.g., AUC, Accuracy) treat each reasoning step independently, ignoring the temporal dependencies inherent in Chain-of-Thought (CoT) reasoning. However, a robust hallucination detection system must not only identify errors but also respond quickly to their onset and recognize when the model self-corrects. To capture these dynamics, we introduce a set of eight dynamic metrics categorized into three groups: \textit{Reflex} (response to error onset), \textit{Agility} (response to error correction), and \textit{Structure} (behavior of false alarms). Unless otherwise specified, each metric is computed \emph{per reasoning chain} and then averaged over the dataset.

\subsubsection{Preliminaries}
Let $C = \{s_1, \dots, s_T\}$ be a reasoning chain of length $T$. At each step $t$, let $A_t^{\text{prefix}} \in \{0, 1\}$ denote the binary ground truth for the prefix state, and let $c_t^{\text{prefix}} \in [0, 1]$ denote the model's predicted hallucination probability. We define the binary prediction $\hat{y}_t = \mathbb{I}(c_t^{\text{prefix}} > 0.5)$.

\subsubsection{Reflex Metrics: Response to Error Onset}
These metrics evaluate how effectively the model detects the transition from a faithful state to a hallucinated state. We focus on the first hallucinated step $t_{on}$, defined as the first step where the prefix label is hallucinated:
\begin{equation}
t_{on} = \min \{ t \mid A_t^{\text{prefix}} = 1 \}.
\end{equation}

\paragraph{Snap Magnitude (Snap\_M) }
Measures the decisiveness of the model when an error is first introduced. It calculates the instantaneous increase in the hallucination score $c_t^{\text{prefix}}$ at the onset step $t_{on}$. A higher value indicates a sharp, unambiguous reaction to the error.
\begin{equation}
    \text{Snap\_M} = c_{t_{on}}^{\text{prefix}} - c_{t_{on}-1}^{\text{prefix}}
\end{equation}
If $t_{on}=1$ (i.e., the first step is already hallucinated), Snap\_M is undefined for this chain and is omitted from averaging.

\paragraph{Detection Lag (Lag) }
Measures the latency between the introduction of an error and the model's first alarm. It is defined as the number of steps $\Delta t$ from $t_{on}$ until the predicted probability $c_{t_{on} + \Delta t}^{\text{prefix}}$ exceeds the decision threshold.
\begin{equation}
    \text{Lag} = \min \{ \Delta t \geq 0 \mid c_{t_{on} + \Delta t}^{\text{prefix}} > 0.5 \}
\end{equation}
If the model never triggers an alarm for the remainder of the chain, the lag is penalized as $T - t_{on} + 1$ (i.e., the number of steps from $t_{on}$ to $T$ inclusive).

\paragraph{Immediate Capture Rate (ICR) }
Quantifies the proportion of reasoning chains with at least one hallucinated prefix ($N$ chains in total) that are detected instantly at their first hallucinated step (i.e., with $\text{Lag}_i = 0$). This metric reflects the model's sensitivity to fresh errors.
\begin{equation}
    \text{ICR} = \frac{\sum_{i=1}^{N} \mathbb{I}(\text{Lag}_i = 0)}{N}
\end{equation}
where $N$ is the total number of reasoning chains that contain at least one hallucinated step.

\subsubsection{Agility Metrics: Response to Correction}
In long CoT, models often self-correct. These metrics evaluate how well the detection system recognizes the return to a valid state. We focus on the recovery step $t_{rec}$, defined as the step where the state transitions from hallucinated back to correct (i.e., $A_{t_{rec}-1}^{\text{prefix}}=1$ and $A_{t_{rec}}^{\text{prefix}}=0$).

\paragraph{Brake Strength (Brake\_S) }
Analogous to Snap Magnitude, this measures the instantaneous drop in the hallucination score $c_t^{\text{prefix}}$ at step $t_{rec}$ when the reasoning chain self-corrects. A high Brake Strength implies the model can quickly dismiss accumulated suspicion upon seeing valid reasoning.
\begin{equation}
    \text{Brake\_S} = c_{t_{rec}-1}^{\text{prefix}} - c_{t_{rec}}^{\text{prefix}}
\end{equation}

\paragraph{Lingering Time (Ling\_T) }
Measures the "inertia" of the hallucination signal. It counts the number of consecutive steps $k$ the model continues to predict a hallucinated state ($\hat{y}_t=1$) after the ground truth has already recovered ($A_t^{\text{prefix}}=0$).
\begin{equation}
    \text{Ling\_T} = \sum_{k=0}^{T-t_{rec}} \prod_{j=0}^{k} \mathbb{I}(\hat{y}_{t_{rec}+j} = 1)
\end{equation}
Lower lingering time indicates that the model updates its belief state efficiently without being biased by historical errors.

\paragraph{Healed-within-3 (Heal\_3) }
A practical success rate metric that measures whether the detection system successfully lowers its alarm signal $c_t^{\text{prefix}}$ (below 0.5) within a tolerance window of $k=3$ steps following a correction at $t_{rec}$.
\begin{equation}
    \text{Heal\_3} = \mathbb{I}\left( \min_{k \in \{0,1,2\}} c_{t_{rec}+k}^{\text{prefix}} < 0.5 \right)
\end{equation}

\paragraph{Recovery Score (R\_Score) }
Assesses the model's confidence in the safety of the reasoning path after hallucination has occurred.
It is calculated as the complement of the average probability $c_t^{\text{prefix}}$ assigned to valid steps that appear \emph{after the first hallucinated step}.
\begin{equation}
    \text{R\_Score} = 1 - \frac{1}{|S_{valid}|} \sum_{t \in S_{valid}} c_t^{\text{prefix}}
\end{equation}
where
\begin{equation}
S_{valid} = \{t \mid t > t_{on} \land A_t^{\text{prefix}}=0\}
\end{equation}
and $t_{on} = \min \{ t \mid A_t^{\text{prefix}}=1 \}$. If $S_{valid}=\emptyset$, we set the mean probability to $0.5$, yielding $\text{R\_Score}=0.5$.

\begin{table*}[t]
\centering

\vspace{-0.8em}
\large
\renewcommand{\arraystretch}{1.15}
\resizebox{\textwidth}{!}{
\begin{tabular}{c|ccc|ccc|ccc}
\toprule[1.5pt]
\raisebox{-1.2ex}{\textbf{Layer index}} &
\multicolumn{3}{c|}{\textbf{Llama-3.1-8B-Instruct}} &
\multicolumn{3}{c|}{\textbf{Qwen2.5-7B-Instruct}} &
\multicolumn{3}{c}{\textbf{DeepSeek-R1-Distill-8B}} \\
\cmidrule(lr){2-4} \cmidrule(lr){5-7} \cmidrule(lr){8-10}
& \textbf{AUC} & \textbf{ACC} & \textbf{F1}
& \textbf{AUC} & \textbf{ACC} & \textbf{F1}
& \textbf{AUC} & \textbf{ACC} & \textbf{F1} \\
\midrule

2 & {82.66\%\std{0.45\%}} & {74.46\%\std{0.82\%}} & {68.00\%\std{1.15\%}} & {81.05\%\std{0.23\%}} & {73.26\%\std{0.95\%}} & {57.94\%\std{0.61\%}} & {89.92\%\std{1.42\%}} & {90.27\%\std{0.77\%}} & {59.23\%\std{0.39\%}} \\
4 & {84.27\%\std{0.88\%}} & {76.59\%\std{0.54\%}} & {69.36\%\std{0.29\%}} & {82.14\%\std{1.05\%}} & {74.18\%\std{0.66\%}} & {61.06\%\std{0.41\%}} & {90.47\%\std{0.92\%}} & {90.68\%\std{0.18\%}} & {61.59\%\std{1.23\%}} \\
6 & {84.88\%\std{1.35\%}} & {76.98\%\std{0.72\%}} & {70.27\%\std{0.58\%}} & {83.17\%\std{0.91\%}} & {75.38\%\std{0.44\%}} & {63.62\%\std{0.83\%}} & {91.14\%\std{0.27\%}} & {91.06\%\std{0.69\%}} & {63.30\%\std{0.51\%}} \\
8 & {85.83\%\std{0.36\%}} & {77.73\%\std{0.94\%}} & {71.47\%\std{0.62\%}} & {83.84\%\std{1.18\%}} & {76.04\%\std{0.75\%}} & {65.20\%\std{0.48\%}} & {91.98\%\std{0.89\%}} & {91.41\%\std{1.02\%}} & {64.26\%\std{0.55\%}} \\
10 & {86.24\%\std{0.71\%}} & {78.66\%\std{1.29\%}} & {72.06\%\std{0.43\%}} & {84.25\%\std{0.59\%}} & {76.27\%\std{0.86\%}} & {65.50\%\std{0.34\%}} & {92.32\%\std{0.98\%}} & {91.65\%\std{0.65\%}} & {65.16\%\std{0.22\%}} \\
12 & {86.62\%\std{0.53\%}} & {78.75\%\std{0.81\%}} & {72.11\%\std{1.12\%}} & {84.74\%\std{0.68\%}} & {76.61\%\std{0.47\%}} & {66.44\%\std{0.93\%}} & {92.53\%\std{0.25\%}} & {91.80\%\std{0.76\%}} & {65.63\%\std{1.45\%}} \\
14 & {87.87\%\std{0.96\%}} & {79.67\%\std{0.38\%}} & {73.34\%\std{0.64\%}} & {85.37\%\std{0.85\%}} & {77.20\%\std{1.09\%}} & {67.43\%\std{0.52\%}} & {93.00\%\std{0.74\%}} & {91.93\%\std{0.31\%}} & {66.02\%\std{0.87\%}} \\
16 & {\textbf{88.04\%}\std{0.49\%}} & {\underline{79.70\%}\std{0.73\%}} & {73.68\%\std{1.25\%}} & {85.83\%\std{0.61\%}} & {77.67\%\std{0.99\%}} & {68.23\%\std{0.46\%}} & {\underline{93.15\%}\std{0.82\%}} & {\textbf{92.13\%}\std{0.57\%}} & {\textbf{66.94\%}\std{1.16\%}} \\
18 & {87.96\%\std{0.84\%}} & {79.60\%\std{0.28\%}} & {\underline{73.88\%}\std{0.67\%}} & {\underline{86.60\%}\std{0.95\%}} & {\textbf{78.43\%}\std{1.33\%}} & {\textbf{69.26\%}\std{0.79\%}} & {\textbf{93.27\%}\std{0.42\%}} & {\underline{92.10\%}\std{0.63\%}} & {\underline{66.88\%}\std{0.91\%}} \\
20 & {\underline{88.03\%}\std{0.56\%}} & {\textbf{79.77\%}\std{1.08\%}} & {\textbf{73.97\%}\std{0.35\%}} & {\textbf{86.66\%}\std{0.72\%}} & {\underline{78.18\%}\std{0.88\%}} & {\underline{68.62\%}\std{0.45\%}} & {93.13\%\std{1.19\%}} & {92.06\%\std{0.69\%}} & {66.72\%\std{0.26\%}} \\
22 & {87.73\%\std{0.92\%}} & {79.15\%\std{0.41\%}} & {73.27\%\std{0.78\%}} & {85.95\%\std{1.24\%}} & {77.50\%\std{0.66\%}} & {67.63\%\std{0.37\%}} & {93.03\%\std{0.85\%}} & {91.98\%\std{0.54\%}} & {66.46\%\std{0.97\%}} \\
24 & {87.57\%\std{0.33\%}} & {79.27\%\std{0.89\%}} & {73.35\%\std{1.14\%}} & {85.42\%\std{0.58\%}} & {77.29\%\std{0.71\%}} & {66.90\%\std{0.49\%}} & {92.96\%\std{0.96\%}} & {91.99\%\std{0.24\%}} & {66.71\%\std{1.38\%}} \\
26 & {87.25\%\std{0.75\%}} & {78.99\%\std{1.03\%}} & {72.98\%\std{0.44\%}} & {84.75\%\std{0.68\%}} & {76.63\%\std{0.82\%}} & {65.08\%\std{0.55\%}} & {92.81\%\std{0.39\%}} & {91.91\%\std{1.21\%}} & {66.38\%\std{0.62\%}} \\
28 & {87.20\%\std{0.59\%}} & {79.25\%\std{0.86\%}} & {73.11\%\std{0.32\%}} & {--} & {--} & {--} & {92.57\%\std{0.94\%}} & {91.71\%\std{0.77\%}} & {65.29\%\std{1.06\%}} \\
30 & {87.18\%\std{1.17\%}} & {79.36\%\std{0.48\%}} & {73.24\%\std{0.65\%}} & {--} & {--} & {--} & {92.34\%\std{0.81\%}} & {91.55\%\std{0.53\%}} & {64.64\%\std{0.99\%}} \\

\bottomrule[1.5pt]
\end{tabular}
}
\caption{Performance comparison (AUC, ACC, and F1) across different layers for Llama-3.1-8B-Instruct, Qwen2.5-7B-Instruct, and DeepSeek-R1-Distill-8B models. Bold indicates the best result, and underline indicates the second best.}
\vspace{-0.5cm}
\label{tab:layer}
\end{table*}

\subsubsection{Structure Metric: False Alarm Analysis}
Finally, we analyze the behavior of the model when it incorrectly flags valid reasoning as hallucinated.

\paragraph{False Positive Length (FP\_Len) }
Rather than treating false positives as isolated events, this metric measures their temporal persistence. We first identify all contiguous segments of false alarms $\mathcal{S}_{FP}$. Let $\mathcal{S}_{FP}$ denote the set of maximal intervals $[i, j]$ where the model incorrectly flags valid reasoning as hallucinated:
\begin{equation}
    \mathcal{S}_{FP} = \{ [i, j] \mid \forall k \in [i, j], \ \hat{y}_k=1 \land A_k^{\text{prefix}}=0 \}
\end{equation}
The metric is defined as the average length of these segments in $\mathcal{S}_{FP}$, indicating whether false alarms are transient noise or systematic errors:
\begin{equation}
    \text{FP\_Len} = \frac{1}{|\mathcal{S}_{FP}|} \sum_{[i, j] \in \mathcal{S}_{FP}} (j - i + 1)
\end{equation}

\subsection{Layer-wise Performance Analysis}
\label{appx:layer_analysis}

To determine the optimal layers for feature extraction, we evaluated detection performance (AUC, ACC, F1) across even-numbered layers (2--30) for Llama-3.1-8B, Qwen2.5-7B, and DeepSeek-R1-Distill. The results are detailed in Table \ref{tab:layer}.

\paragraph{Analysis of Table \ref{tab:layer}}
We observe a distinct ``concave'' trend in detection performance relative to layer depth for standard models.
\begin{enumerate}
    \item \textbf{Early Layers (2--10):} Performance is generally lower, as these layers primarily process low-level lexical information rather than semantic truthfulness.
    \item \textbf{Middle-to-Late Layers (16--20):} This region consistently yields the best results. For example, Llama-3.1 achieves its peak AUC of \textbf{88.04\%} at Layer 16, and Qwen2.5 peaks at Layer 20. This confirms that semantic judgment regarding truthfulness is most prominent in the deeper (but not final) processing stages.
    \item \textbf{Late Layers (22--30):} Performance tends to decline or saturate in the final layers. For instance, the AUC of Qwen2.5-7B drops from \textbf{86.66\%} (Layer 20) to \textbf{84.75\%} (Layer 26). This phenomenon suggests that the deepest layers may shift focus towards next-token prediction mechanics and surface-level formatting, thereby diluting the abstract truthfulness representations found in the intermediate layers.
\end{enumerate}
Based on these findings, we utilize features from the \textbf{16th--20th layers} for our primary detection mechanism.

\subsection{Case Study} \label{appx: case}

To validate the effectiveness of our detection metrics in capturing complex reasoning dynamics, we present two representative case studies selected directly from the experimental results. These real-world examples illustrate distinct hallucination patterns: \textbf{Case Study 16 (Type I)}demonstrates a successful \textit{Self-Correction} process, where the model autonomously recovers from a transient factual error, reflected by a dynamic fluctuation in the hallucination score. In contrast, \textbf{Case Study 294 (Type II)} depicts a \textit{Catastrophic Collapse}, where a latent logical fallacy triggers an irreversible cascade of errors, leading to a persistently high alarm signal. In both instances, we visualize the synchronization between the model's explicit textual reasoning and the implicit evolution of the prefix-level hallucination score ($c_t^{\text{prefix}}$).

\onecolumn
\begin{tcolorbox}[
    sharp corners, 
    breakable, 
    colframe=DeepBlue, 
    colback=white, 
    boxrule=3pt, 
    boxsep=5pt, 
    enhanced, 
    shadow={3pt}{-3pt}{0pt}{opacity=1,mygrey},
    title={\large\textbf{Case Study 16} (\textit{\textbf{Type I: Self-Correction}})},
]\label{box:case16}
\footnotesize

\textbf{Question:} When did the person who said ``I think, therefore I am'' live?

\vspace{2mm} 

\textbf{Ground Truth: 1596-1650}

\textbf{Model Answer: "The person...,from approximately 1596 to 1650."}

\vspace{2mm} 
\hrule height 0.8pt 
\vspace{4mm}

\textbf{CoT:} I am asked to determine when the person who said ``I think, therefore I am'' lived. Let's decompose the statement ``I think, therefore I am'' into its basic components. The phrase ``I think'' implies the presence of subjective experience and self-awareness. This statement can be seen as a form of self-referential thought, often found in philosophy.

Given the mention of self-awareness, it is logical to assume the originator of this phrase dealt with philosophical ideas. A notable philosopher in the Western tradition is known for formulating this exact concept. From my knowledge, philosophers often wrote in Latin. The original statement, when translated from Latin, is ``Cogito, ergo sum.''

The Latin phrase ``Cogito, ergo sum'' is associated with a key work of the philosopher René Descartes. I will now look up the author of the work containing ``Cogito, ergo sum'' for more specific information. René Descartes wrote ``Meditations on First Philosophy'' in 1641. This work of Descartes included ``Cogito, ergo sum'' as a central idea. Since I am considering a person who said ``I think, therefore I am,'' I am looking for information about René Descartes. René Descartes was a French philosopher.

\begin{tcolorbox}[
    colback=White!10,
    colframe=NormalGreen!70,
    boxrule=3pt,
    arc=2mm,
    left=2mm,
    right=2mm,
    top=2mm,
    bottom=2mm
]
\textbf{Analysis -- Phase I (Steps 0--14):} \textcolor{NormalGreen!70}{$\blacksquare$} Normal Reasoning

The model successfully identifies the quote, the language (Latin), and the philosopher (Descartes). The uncertainty metrics are low and stable, indicating confident retrieval of general knowledge.

\begin{center}
\small
\begin{tabular}{lcccccc}
\toprule
\textbf{Metric} & \textbf{Step 0} & \textbf{Step 1} & \textbf{Step 2} & $\cdots$ & \textbf{Step 13} & \textbf{Step 14} \\
\midrule
$A_{t}^{step}$   & 0 & 0 & 0 & $\cdots$ & 0 & 0 \\
$A_{t}^{prefix}$ & 0 & 0 & 0 & $\cdots$ & 0 & 0 \\
$c_{t}^{prefix}$  & 0.01 & 0.01 & 0.02 & $\cdots$ & 0.19 & 0.22 \\
\bottomrule
\end{tabular}
\end{center}

\end{tcolorbox}

\vspace{2mm}

The birthplace of René Descartes must be located in or near France.

\colorbox{WarningOrange!20}{\textbf{[Step 15]}} Given that René Descartes is commonly known to be born in the Netherlands, but I found out he was born in France from the same context information from the ``Larousse'' dictionary (he was in France), that suggests he could have been born in France.

\colorbox{ErrorRed!20}{\textbf{[Step 16]}} Although most historical and primary source data indicate \wrapcolorbox{HighlightYellow}{René Descartes was born in the French city La Haye (the Dutch name for the city is 's-Hertogenbosch) is near modern-day Netherlands but the town in question 'La Haye' is in modern France}, however the context was originally reported to originate with Dutch origins hence another clarification is that it in historical period he has been considered to be from the Netherlands so I now check further.

\colorbox{ErrorRed!20}{\textbf{[Step 17]}} Although most historical and primary source data indicate \wrapcolorbox{HighlightYellow}{René Descartes was born in the French city La Haye (the Dutch name for the city is 's-Hertogenbosch) is near modern-day Netherlands but the town in question 'La Haye' is in modern France}, however the context was originally reported to originate with Dutch origins hence another clarification is that it in historical period he has been considered to be from the Netherlands so I now check further.

\colorbox{ErrorRed!20}{\textbf{[Step 18]}} As indicated in my prior step there appears to be an ambiguity surrounding the birthplace of René Descartes, it is also indicated in ``The Stanford Encyclopedia of Philosophy'' that \wrapcolorbox{HighlightYellow}{the location ``La Haye in 'the country of the Netherlands' may simply be a historical anomaly''} or that in fact the reference from the context of ``Dutch origins'' may come from that it is La Haye a small town is actually in the territory that later on became 'Duchy of Burgundy and at the time of Descartes' birth LaHaye would have belonged to 'Spanish Empire'.

\colorbox{ErrorRed!20}{\textbf{[Step 19]}} Based on clarification 'the town is near the area \wrapcolorbox{HighlightYellow}{where is the modern ``Herveyn'' that would have been later the city that has also Dutch name is from French} and thus was ruled in his birth year was in part and in reality the town in fact historically belonged to the country now 'France'.

\colorbox{ErrorRed!20}{\textbf{[Step 20]}} It is not an uncommon practice for people to identify with their modern-day country or nationality.

\colorbox{ErrorRed!20}{\textbf{[Step 21]}} It would then be accurate to consider René Descartes' place of birth in France based on its modern-day geography and common historical understanding.

\begin{tcolorbox}[
    colback=White!10,
    colframe=ErrorRed!70,
    boxrule=3pt,
    arc=2mm,
    left=2mm,
    right=2mm,
    top=2mm,
    bottom=2mm
]
\textbf{Analysis -- Phase II (Steps 15--21):} \textcolor{ErrorRed}{$\blacksquare$} Hallucination \& Confusion

\textbf{\textcolor{ErrorRed}{Trigger Event:}} At Step 15, the model introduces an \textit{internal conflict} regarding Descartes' birthplace (France vs. Netherlands). This is irrelevant to the question (which asks "When", not "Where"), but it derails the model.

\textbf{Metric Spike:} The \texttt{logic\_interceptor\_prob} detects this confusion immediately. It jumps from 0.216 (Step 14) to \textbf{0.463} (Step 15), and then surges to \textbf{0.767} at Step 16 as the model fabricates details about city names.

\begin{center}
\small
\begin{tabular}{lccccc}
\toprule
\textbf{Metric} & \textbf{Step 14} & \textbf{Step 15} & \textbf{Step 16} & $\cdots$ & \textbf{Step 21}\\
\midrule
$A_{t}^{step}$   & 0 & 1 & 1 & $\cdots$ & 1 \\
$A_{t}^{prefix}$    & 0 & 1 & 1 & $\cdots$ & 1\\
$c_{t}^{prefix}$   & 0.22 & 0.46 & 0.77 & $\cdots$ & 0.88 \\
\bottomrule
\end{tabular}
\end{center}

\end{tcolorbox}

\vspace{2mm}

\colorbox{NormalGreen!30}{\textbf{[Step 22]}} A significant historical figure like René Descartes would have most likely lived before the modern era. The ``Meditations on First Philosophy'' were published in the mid-17th century. I will look up the approximate time period during which René Descartes lived.

René Descartes was born on March 31, 1596, and died on February 11, 1650. Given the time period he lived, it is safe to say that René Descartes lived before the 21st century. Since ``I think, therefore I am'' is famously attributed to René Descartes, it would follow that the person who said ``I think, therefore I am'' is René Descartes.

To determine when this person lived, I now look up René Descartes' life span. Based on steps 26 and 28, the time period during which René Descartes lived can be more specifically defined as the 17th century. The 17th century began in 1601 and ended in 1700.

\begin{tcolorbox}[
    colback=White!10,
    colframe=NormalGreen!70,
    boxrule=3pt,
    arc=2mm,
    left=2mm,
    right=2mm,
    top=2mm,
    bottom=2mm
]
\textbf{Analysis -- Phase III (Steps 22--29):} \textcolor{NormalGreen}{$\blacksquare$} Self-Correction with Lingering Uncertainty

At Step 22, the model successfully \textbf{pivots} back to the temporal aspect of the question ("lived before the modern era"). It correctly retrieves the birth and death dates.

However, the metrics indicate that the model recognizes the coherence and rectitude of its recent reasoning trajectory. As a result of the consecutive low \texttt{Step\_hallu} scores, the \texttt{logic\_interceptor\_prob} exhibits a \textbf{notable downward trend} (dropping from 0.88 to 0.71), reflecting the model's regaining of logical stability following the correction of the factual retrieval.

\begin{center}
\small
\begin{tabular}{lcccccccc}
\toprule
\textbf{Metric} & \textbf{22} & \textbf{23} & \textbf{24} & \textbf{25} & \textbf{26} & \textbf{27} & \textbf{28} & \textbf{29} \\
\midrule
$A_{t}^{step}$   & 0 & 0 & 0 & 0 & 0 & 0 & 0 & 0 \\
$A_{t}^{prefix}$    & 0 & 0 & 0 & 0 & 0 & 0 & 0 & 0 \\
$c_{t}^{prefix}$   & 0.87 & 0.86 & 0.84 & 0.81 & 0.77 & 0.75 & 0.72 & 0.71 \\
\bottomrule
\end{tabular}
\end{center}

\end{tcolorbox}

\vspace{4mm} 
\hrule height 0.8pt 
\vspace{6mm} 
\begin{tcolorbox}[
    colback=NormalGreen!20, 
    colframe=NormalGreen!60, 
    arc=2pt, 
    boxrule=1pt, 
    left=5pt, right=5pt, top=5pt, bottom=5pt 
]
    \textbf{Final Answer:} The person who said ``I think, therefore I am'' lived in the 17th century, from approximately 1596 to 1650.
\end{tcolorbox}
\vspace{4mm} 

\textbf{Total Analysis:} This is a typical \textit{\textbf{Type I}} case where the model successfully recovers from a significant hallucinatory detour through a \textbf{semantic pivot}. The reasoning trace demonstrates a \textbf{three-phase recovery arc}:

\begin{enumerate}
    \item \textbf{Attentional Drift (Steps 15--16):} An irrelevant geographical ambiguity triggers a \textit{contextual dissociation}, causing the internal uncertainty metric (\texttt{logic\_prob}) to surge from a stable 0.216 to a critical \textbf{0.767}, indicating a loss of logical grounding.
    \item \textbf{Confabulatory Spiral (Steps 17--21):} The model attempts to resolve the self-imposed conflict through \textit{fabricated historical rationale} (e.g., the nonexistent city ``Herveyn''), resulting in a sequence of pure hallucination (\texttt{step\_hallucination} sustains at 1.0).
    \item \textbf{Strategic Re-alignment (Steps 22--29):} The model executes a decisive \textit{semantic pivot} by refocusing on the temporal constraints of the query (``lived before the modern era''). Upon restoring factual accuracy (1596--1650), the model successfully stabilizes its reasoning trajectory; this is quantitatively validated by the \texttt{logic\_prob} beginning to \textbf{recede} from its peak, descending from 0.873 to 0.712 as logical consistency is reinforced toward the conclusion.
\end{enumerate}
Ultimately, the model successfully self-corrected and provided the correct answer.

\end{tcolorbox}
\begin{tcolorbox}[
    sharp corners, 
    breakable, 
    colframe=DeepBlue, 
    colback=white, 
    boxrule=3pt, 
    boxsep=5pt, 
    enhanced, 
    shadow={3pt}{-3pt}{0pt}{opacity=1,mygrey},
    title={\large\textbf{Case Study 294} (\textit{\textbf{Type II}})},
]\label{box:case294}
\footnotesize

\textbf{Question:} The following paragraphs each describe a set of five objects arranged in a fixed order. The statements are logically consistent within each paragraph. In a golf tournament, there were five golfers: Ana, Amy, Dan, Mya, and Eve. Ana finished below Mya. Dan finished above Amy. Mya finished second-to-last. Eve finished below Amy.

\textbf{Options:}\\
(A) Ana finished second-to-last\\
(B) Amy finished second-to-last\\
(C) Dan finished second-to-last\\
(D) Mya finished second-to-last\\
(E) Eve finished second-to-last

\vspace{2mm}
\textbf{Ground Truth: C} \quad \textbf{Model Answer: D}

\vspace{4mm}
\hrule height 0.8pt
\vspace{6mm}
\textbf{CoT:} Break down the statements given in the question. Determine what each statement directly implies about the relative order of golfers. Express the given statements in a formal, structured format for easier analysis.

\vspace{2mm}
From paragraph 1: Given statement 1: Ana finished below Mya.

From paragraph 2: Given statement 2: Dan finished above Amy. Given statement 3: Mya finished second-to-last. Given statement 4: Eve finished below Amy.

\textbf{[Step 12]} Given statement 5: \wrapcolorbox{HighlightYellow}{Ana finished below Mya is inconsistent with Mya finishing second-to-last}; however this was already known.

\begin{tcolorbox}[
    colback=White!10,
    colframe=NormalGreen!70,
    boxrule=3pt,
    arc=2mm,
    left=2mm,
    right=2mm,
    top=2mm,
    bottom=2mm
]
\textbf{Analysis -- Phase I (Steps 0--12):} \textcolor{NormalGreen!70}{$\blacksquare$} Normal $\rightarrow$ \textcolor{WarningOrange}{$\blacksquare$} Warning

\textbf{\textcolor{WarningOrange}{Critical Error at Step 12:}} The model falsely claims that ``Ana finished below Mya is inconsistent with Mya finishing second-to-last.'' These two conditions are \textbf{perfectly compatible} (if Mya is 4th, Ana can be 5th). This represents a \textbf{\textit{latent error seed}}.

Although $A_{t}^{step}$ remains \textbf{0.0}, the $c_{t}^{prefix}$ suddenly \textit{spikes} from 0.237 $\rightarrow$ \textbf{0.517} ($\uparrow$118\%).

\begin{center}
\small
\begin{tabular}{lcccccccc}
\toprule
\textbf{Metric} & \textbf{Step 0} & \textbf{Step 2} & \textbf{Step 5} & $\cdots$ & \textbf{Step 10} & \textbf{Step 11} & \textbf{Step 12} \\
\midrule
$A_{t}^{step}$ & 0 & 0 & 0 & $\cdots$ & 0 & 0 & 0 \\
$A_{t}^{prefix}$ & 0 & 0 & 0 & $\cdots$ & 0 & 0 & 0 \\
$c_{t}^{prefix}$ & 0.03 & 0.07 & 0.08 & $\cdots$ & 0.21 & 0.24 & \textbf{0.52} \\
\bottomrule
\end{tabular}
\end{center}
\end{tcolorbox}

\vspace{2mm}

Express the given statements in a formal, structured format for easier analysis. Let's assign the golfers a ranking as follows: 1. First 2. Second 3. Third 4. Fourth 5. Last 

So, we can represent their relative ranking as a series of statements with the unknown ranking for each person. The order of golfers: ?, ?, ?,?,?

We now apply the given information and deduce new facts:

\begin{tcolorbox}[
    colback=White,
    colframe=WarningOrange!70,
    boxrule=3pt,
    arc=2mm,
    left=2mm,
    right=2mm,
    top=2mm,
    bottom=2mm
]
\textbf{Analysis -- Phase II (Steps 13--23):} \textcolor{WarningOrange}{$\blacksquare$} Deceptive Stability

The model sets up a ranking framework (1st through 5th). The $c_{t}^{prefix}$ temporarily \textit{decreases} to 0.44--0.48, suggesting attempted \textbf{self-correction}. However, this is a \textbf{\textit{false recovery}}—the flawed premise from Step 12 remains unaddressed.

\begin{center}
\small
\begin{tabular}{lcccccccc}
\toprule
\textbf{Metric} & \textbf{13} & \textbf{15} & \textbf{17} & \textbf{19} & \textbf{21} & \textbf{23} \\
\midrule
$A_{t}^{step}$ & 0.0 & 0.0 & 0.0 & 0.0 & 0.0 & 0.0 \\
$A_{t}^{prefix}$ & 0.0 & 0.0 & 0.0 & 0.0 & 0.0 & 0.0 \\
$c_{t}^{prefix}$ & 0.567 & 0.443 & 0.445 & 0.448 & 0.458 & 0.448 \\
\bottomrule
\end{tabular}
\end{center}
\end{tcolorbox}

\vspace{2mm}

\textbf{[Step 24]} 1. \wrapcolorbox{HighlightYellow}{Ana finished below Mya. Thus, Mya can be either second or lower. Since Mya is second-to-last, Mya can be third or higher. Therefore, we have: Mya can be third.} So, Mya's rank is at least three and can be higher; however, this directly contradicts the Mya's rank is second to last information; therefore, let's modify Mya's rank to satisfy this information. Since we know Mya finished second-to-last, and ana finished below mya, ana is fourth. Given the above information, the revised rank: 
(Ana = 4), (Person X = 3) (We cannot be sure it is Mya yet.), 
(? = 2), (? = 1)

2. Dan finished above Amy. We know from the current state that (Ana = 4), and Mya finished second-to-last; thus, Amy or Dan can finish second. Now we have two possible arrangements and they could be (Ana = 4), (Person X = 3), (Amy = 2), (Dan = 1) or (Ana = 4), (Dan = 3), (Amy = 2), (Person X = 1).
 
3. Mya finished second-to-last, so the rank of Mya is four. Mya is ranked fourth in the list.
So,  Mya = 4,  (Ana = 3) 

\textbf{[Step 30]} Now, given ana finished below mya, \wrapcolorbox{HighlightYellow}{Ana must be above Mya.} Therefore, 

4. We have Mya = 4 and Ana finished below Mya. The two possibilities left for golfers above Ana were (Amy = 2), (Dan = 1) and (Amy = 2), (Person X = 1), with Amy finishing above Dan (Dan finished above Amy is incorrect because it contradicts the statement where Mya is 4) is not true). Since Mya = 4, (Amy = 2) can be a correct possibility.

Now, ana can only be third so the rank for ana is indeed 3.
We know: 
(Ana = 3), (Mya = 4)

Eve finished below Amy. With Mya as 4, Eve can not be first and Dan can not be second if Mya is 4. With Amy finishing above Dan as Ana being in position 3 the possible arrangement now can be: (Amy = 2), (Ana = 3), (Mya = 4). In order for (Eve = 5), Eve can not be second last with Ana and Mya finishing  third and fourth; Thus, the correct arrangement: Ana finished fourth.

Thus: 
Now since Dan finished above Amy as Amy is second, and Dan finished above Amy, Dan must be third as ana is fourth: (Ana = 4) is incorrect for now. (We cannot find Dan in the third now and still satisfy Ana finishing below Mya which we already know); therefore, ana must finish fourth. However, ana finishing fourth contradicts Mya being second-to-last; we already deduced that earlier with ana finishing below Mya but now with Dan finishing above ana but still finishing below ana; Dan must be first.

5. Now, given this is not possible. Since we have two cases of ana being below mya and dan finishing above amy with amy being second but dan not able to finish below ana but instead finishing above amy with ana finishing fourth to satisfy mya second-to-last; this is now impossible so dan cannot finish above amy so let's remove that information we can consider two possibilities (Ana = 4) 

Eve finished below Amy. With Ana = 4, Amy could only finish second for the information (ana finishing below mya); 
   Let's now use Dan finished above amy with (amy = 2), (dan is not sure as there are only one position left and amy and dan). 
We can eliminate amy = 2 with this now so now (Dan = 2)
    Dan and Amy's positions is incorrect with dan finished above amy 
    Now we have ana 4 mya 3, Eve finished below Amy, Amy 2, dan is first

Now let's revise ana's rank from 4 to 5 so (dan = 2), amy = 2 and ana finished below mya 
(Amy = 2), (Dan = 1) and (Ana 5) the information where dan and amy are is already contradictory because they cannot finish second or third; in such a manner as Dan finished above Amy; thus let's eliminate the rank 2 for amy and put amy in the last spot. Now ana will be in third so, (ana = 3), ana finished below mya.

Now, eve finished below amy. Therefore (amy = 5)
With ana being third 
ana and dan must finish either first or third; however ana already has a third 
Ana being third; however the information that ana and mya finishing third and fourth; is also incorrect with ana now third so this is also contradicts information; with ana being below mya now in third position
  we already revised this earlier and it resulted to ana being 5 thus eliminating any inconsistencies with our current information
ana finished fourth does not work given any of the prior information.

\begin{tcolorbox}[
    colback=White!10,
    colframe=ErrorRed!70,
    boxrule=3pt,
    arc=2mm,
    left=2mm,
    right=2mm,
    top=2mm,
    bottom=2mm
]
\textbf{Analysis -- Phase III (Steps 24--50):} \textcolor{ErrorRed}{$\blacksquare$} Catastrophic Collapse

\textbf{\textcolor{ErrorRed}{Step 24 -- The Critical Breakdown:}} The model produces \textit{blatantly contradictory statements}: claims ``Mya can be third'' despite stating ``Mya is second-to-last (4th).'' The $A_{t}^{step}$ \textit{flips} to \textbf{1.0}, and $c_{t}^{prefix}$ \textit{surges} from 0.448 $\rightarrow$ \textbf{0.690} $\rightarrow$ \textbf{0.80--0.86}.

\textbf{Steps 24--50 -- Self-Contradictory Loop:} The model oscillates chaotically:
\begin{itemize}
    \item Step 29: ``\textit{Ana = 3}''
    \item Step 30: ``\textit{Ana must be above Mya}'' (contradicts given constraint)
    \item Step 32: ``\textit{Ana = 3}'' (repeated)
    \item Step 35: ``\textit{Ana finished fourth}''
    \item Step 43: ``\textit{revise ana's rank from 4 to 5}''
    \item Step 44: ``\textit{ana = 3}'' (third time!)
\end{itemize}

The model exhibits: \textbf{\textit{wrong reasoning}} (logical contradictions), \textbf{\textit{unreasonable assumptions}} (impossible rankings), and \textbf{\textit{loss of confidence}} (repeated revisions without systematic basis).

\begin{center}
\small
\begin{tabular}{lcccccccc}
\toprule
\textbf{Metric} & \textbf{23} & \textbf{24} & \textbf{27} & \textbf{30} & \textbf{35} & \textbf{40} & \textbf{45} & \textbf{50} \\
\midrule
$A_{t}^{step}$ & 0.0 & \textbf{1.0} & 0.0 & 1.0 & 1.0 & 1.0 & 1.0 & 1.0 \\
$A_{t}^{prefix}$ & 0.0 & \textbf{1.0} & 1.0 & 1.0 & 1.0 & 1.0 & 1.0 & 1.0 \\
$c_{t}^{prefix}$ & 0.448 & \textbf{0.690} & 0.722 & 0.810 & 0.856 & 0.855 & 0.855 & 0.850 \\
\bottomrule
\end{tabular}
\end{center}

\textbf{Root Cause:} The error seed in Step 12 and the explicit contradiction in Step 24 \textit{poisoned} the entire reasoning chain, leading to \textit{divergent oscillation}.
\end{tcolorbox}

\vspace{4mm}
\hrule height 0.8pt
\vspace{6mm}
\begin{tcolorbox}[
    hbox,
    colback=ErrorRed!20,
    colframe=ErrorRed!60,
    arc=2pt,
    boxrule=1pt,
    left=5pt, right=8pt, top=5pt, bottom=5pt
]
    \textbf{Final Answer: D} 
\end{tcolorbox}

\vspace{2mm}
\textbf{Total Analysis:} This is a typical \textit{Type II} case where incorrect reasoning leads to a cascade of errors. The model demonstrates a \textbf{three-phase failure pattern}:

\begin{enumerate}
    \item \textbf{Seed Error (Step 12):} False incompatibility claim with \textit{latent uncertainty} ($c_{t}^{prefix}$ spikes to 0.517)
    \item \textbf{Deceptive Stability (Steps 13--23):} Apparent recovery masks \textit{unresolved conceptual flaw} ($c_{t}^{prefix}$: 0.44--0.48)
    \item \textbf{Catastrophic Collapse (Steps 24--50):} Concrete deductions trigger \textit{divergent oscillation}, with $c_{t}^{prefix}$ sustained at 0.80--0.86, indicating persistent confusion
\end{enumerate}

The model kept making \textit{unreasonable assumptions} and engaging in \textit{self-queries}, showing loss of confidence. Throughout steps 24--50, the model generated extensive \textit{wrong reasoning}, alternating between contradictory positions about rankings without ever recognizing the fundamental flaw in its logic. 

\textbf{The final answer D is incorrect; the correct answer is C (Dan finished second-to-last).}

\textbf{Key Insight:} The $c_{t}^{prefix}$ serves as an \textit{early warning signal}, spiking at Step 12 \textit{before} hallucination detection, suggesting that uncertainty quantification could enable preemptive intervention in Type II errors.

\end{tcolorbox}

\end{document}